\documentclass[12pt, onecolumn, draftclsnofoot, journal]{IEEEtran}

\usepackage[encapsulated]{CJK}
\usepackage{ucs}
\usepackage[utf8x]{inputenc}
\usepackage[cmex10]{amsmath}
\usepackage{amssymb,amscd,bbm,amsthm,mathrsfs,dsfont}
\usepackage{algorithmic,algorithm}
\usepackage{mdwmath}
\usepackage{mdwtab}
\usepackage{bm,upgreek}
\usepackage{cite}
\usepackage{graphicx,psfrag}
\usepackage{array}
\usepackage{booktabs}
\usepackage{indentfirst}
\usepackage{subfigure}
\usepackage{lipsum,fancyhdr,lastpage,refcount}
\usepackage{mathtools}
\usepackage[T1]{fontenc}
\usepackage{url}

\IEEEoverridecommandlockouts

\let\oldnl\nl
\newcommand{\nonl}{\renewcommand{\nl}{\let\nl\oldnl}}

\begin{document}

\title{Optimized Computation Offloading Performance in Virtual Edge Computing Systems via Deep Reinforcement Learning}

\author{\IEEEauthorblockN{Xianfu Chen, Honggang Zhang, Celimuge Wu, Shiwen Mao, Yusheng Ji, and Mehdi Bennis}

\thanks{X. Chen is with the VTT Technical Research Centre of Finland, Finland (e-mail: xianfu.chen@vtt.fi). H. Zhang is with the College of Information Science and Electronic Engineering, Zhejiang University, Hangzhou, China (e-mail: honggangzhang@zju.edu.cn). C. Wu is with the Graduate School of Informatics and Engineering, University of Electro-Communications, Tokyo, Japan (email: clmg@is.uec.ac.jp). S. Mao is with the Department of Electrical and Computer Engineering, Auburn University, Auburn, AL, USA (email: smao@ieee.org). Y. Ji is with the Information Systems Architecture Research Division, National Institute of Informatics, Tokyo, Japan (e-mail: kei@nii.ac.jp). M. Bennis is with the Centre for Wireless Communications, University of Oulu, Finland (email: bennis@ee.oulu.fi).}
}

\maketitle

\begin{abstract}

To improve the quality of computation experience for mobile devices, mobile-edge computing (MEC) is a promising paradigm by providing computing capabilities in close proximity within a sliced radio access network (RAN), which supports both traditional communication and MEC services.
Nevertheless, the design of computation offloading policies for a virtual MEC system remains challenging.
Specifically, whether to execute a computation task at the mobile device or to offload it for MEC server execution should adapt to the time-varying network dynamics.
In this paper, we consider MEC for a representative mobile user in an ultra-dense sliced RAN, where multiple base stations (BSs) are available to be selected for computation offloading.
The problem of solving an optimal computation offloading policy is modelled as a Markov decision process, where our objective is to maximize the long-term utility performance whereby an offloading decision is made based on the task queue state, the energy queue state as well as the channel qualities between MU and BSs.
To break the curse of high dimensionality in state space, we first propose a double deep $Q$-network (DQN) based strategic computation offloading algorithm to learn the optimal policy without knowing a priori knowledge of network dynamics.
Then motivated by the additive structure of the utility function, a $Q$-function decomposition technique is combined with the double DQN, which leads to novel learning algorithm for the solving of stochastic computation offloading.
Numerical experiments show that our proposed learning algorithms achieve a significant improvement in computation offloading performance compared with the baseline policies.

\end{abstract}


\section{Introduction}
\label{intr}

With the proliferation of smart mobile devices, a multitude of mobile applications are emerging and gaining popularity, such as location-based virtual/augmented reality and online gaming \cite{Cisc17}.
However, mobile devices are in general resource-constrained, for example, the battery capacity and the local CPU computation power are limited.
When executed at the mobile devices, the performance and Quality-of-Experience (QoE) of computation-intensive applications are significantly affected by the devices' limited computation capabilities.
The tension between computation-intensive applications and resource-constrained mobile devices creates a bottleneck for having a satisfactory Quality-of-Service (QoS) and QoE, and is hence driving a revolution in computing infrastructure \cite{Saty17}.

In contrast to cloud computing, mobile-edge computing (MEC) is envisioned as a promising paradigm, which provides computing capabilities within the radio access networks (RANs) in close proximity to mobile users (MUs) \cite{Mach17}.
By offloading computation tasks to the resource-rich MEC servers, not only the computation QoS and QoE can be greatly improved, but the capabilities of mobile devices can be augmented for running a variety of resource-demanding applications.
Recently, lots of efforts have been put to the design of computation offloading policies.
In \cite{Wang17}, Wang et al. developed an alternating direction method of multipliers-based algorithm to solve the problem of revenue maximization by optimizing computation offloading decision, resource allocation and content caching strategy.
In \cite{Hu18}, Hu et al. proposed a two-phase based method for joint power and time allocation when considering cooperative computation offloading in a wireless power transfer-assisted MEC system.
In \cite{Wang18}, Wang et al. leveraged a Lagrangian duality method to minimize the total energy consumption in a computation latency constrained wireless powered multiuser MEC system.

For a MEC system, the computation offloading requires wireless data transmission, hence how to allocate wireless radio resource between the traditional communication service and the MEC service over a common RAN raises a series of technical challenges.
Network slicing is a key enabler for RAN sharing, with which the traditional single ownership of network infrastructure and spectrum resources can be decoupled from the wireless services \cite{Cost13}.
Consequently, the same physical network infrastructure is able to host multiple wireless service providers (WSPs) \cite{Kokk12, Chen1801}.
In literature, there exist several efforts investigating joint communication and computation resource management in such virtualized networks, which support both the traditional communication service and the MEC service \cite{Zhao18, Zhou17}.
In this work, we focus on designing optimal stochastic computation offloading policies in a sliced RAN, where a centralized network controller (CNC) is responsible for control-plane decisions on wireless radio resource orchestration over the traditional communication and MEC services.

The computation offloading policy designs in previous works \cite{Wang17, Hu18, Zhao18, Zhou17, You17, Lyu18} are mostly based on one-shot optimization and fail to characterize long-term computation offloading performance.
In a virtual MEC system, the design of computation offloading policies should account for the environmental dynamics, such as the time-varying channel quality and the task arrival and energy status at a mobile device.
In \cite{Liu16}, Liu et al. formulated the problem of delay-optimal computation task offloading under a Markov decision process (MDP) framework and developed an efficient one-dimensional search algorithm to find the optimal solution.
However, the challenge lies in the dependence on statistical information of channel quality variations and computation task arrivals.
In \cite{Mao16}, Mao et al. investigated a dynamic computation offloading policy for a MEC system with wireless energy harvesting-enabled mobile devices using a Lyapunov optimization technique.
The same technique was adopted to study the power-delay tradeoff in the scenario of computation task offloading by Liu et al. \cite{Liu17} and Jiang et al. \cite{Jian15}.
The Lyapunov optimization can only construct an approximately optimal solution.
Xu et al. developed in \cite{Xu17} a reinforcement learning based algorithm to learn the optimal computation offloading policy, which at the same time does not need a priori knowledge of network statistics.

When the MEC meets an ultra dense sliced RAN, multiple base stations (BSs) with different data transmission qualities are available for offloading a computation task.
In this context, the explosion in state space makes the conventional reinforcement learning algorithms \cite{Xu17, Watk12, Rich98} infeasible.
Moreover, in this paper, wireless charging \cite{Alsh17} is integrated into a MEC system, which on one hand achieves sustained computation performance but, on the other hand, makes the design of a stochastic computation offloading policy even more challenging.
The main contributions in this work are four-fold.
Firstly, we formulate the stochastic computation offloading problem in a sliced RAN as a MDP, in which the time-varying communication qualities and computation resources are taken into account.
Secondly, to deal with the curse of state space explosion, we resort to a deep neural network based function approximator \cite{Mnih15} and derive a double deep $Q$-network (DQN) \cite{Hass16} based reinforcement learning (DARLING) algorithm to learn the optimal computation offloading policy without any a priori knowledge of network dynamics.
Thirdly, by further exploring the additive structure of the utility function, we attain a novel online deep state-action-reward-state-action based reinforcement learning algorithm (Deep-SARL) for the problem of stochastic computation offloading.
To the best knowledge of the authors, this is the first work to combine a $Q$-function decomposition technique with the double DQN.
Finally, numerical experiments based on TensorFlow are conducted to verify the theoretical studies in this paper.
It shows that both of our proposed online learning algorithms outperform three baseline schemes.
Especially, the Deep-SARL algorithm achieves the best computation offloading performance.

The rest of the paper is organized as follows.
In the next section, we describe the system model and the assumptions made throughout this paper.
In Section \ref{prob}, we formulate the problem of designing an optimal stochastic computation offloading policy as a MDP.
We detail the derived online learning algorithms for stochastic computation offloading in a virtual MEC system in Section \ref{solu}.
To validate the proposed studies, we provide numerical experiments under various settings in Section \ref{simu}.
Finally, we draw the conclusions in Section \ref{conc}.
In Table \ref{tabl1}, we summarize the major notations of this paper.
\begin{table}[t]
  \caption{Major notations used in the paper.}\label{tabl1}
        \begin{center}
        \begin{tabular}{c|l}
              \hline
              $W$                                           & bandwidth of the spectrum allocated for MEC                           \\\hline
              $B$/$\mathcal{B}$                             & number/set of BSs                                                     \\\hline
              $X/\mathcal{X}$                               & number/set of network states                                          \\\hline
              $Y/\mathcal{Y}$                               & number/set of joint control actions                                   \\\hline
              $\delta$                                      & duration of one decision epoch                                        \\\hline
              $g_b$, $g_b^j$                                & channel gain state between the MU and BS $b$                          \\\hline
              $q_{(\mathrm{t})}$, $q_{(\mathrm{t})}^j$      & task queue state of the MU                                            \\\hline
              $q_{(\mathrm{t})}^{(\max)}$                   & maximum task queue length                                             \\\hline
              $\mu$                                         & input data size of a task                                             \\\hline
              $\nu$                                         & required CPU cycles for a task                                        \\\hline
              $f_{(\mathrm{CPU})}^{(\max)}$                 & maximum CPU-cycle frequency                                           \\\hline
              $p_{(\mathrm{tr})}^{(\max)}$                  & maximum transmit power                                                \\\hline
              $a_{(\mathrm{t})}$, $a_{(\mathrm{t})}^j$      & computation task arrival                                              \\\hline
              $q_{(\mathrm{e})}$, $q_{(\mathrm{e})}^j$      & energy queue state of the MU                                          \\\hline
              $q_{(\mathrm{e})}^{(\max)}$                   & maximum energy queue length                                           \\\hline
              $a_{(\mathrm{e})}$, $a_{(\mathrm{e})}^j$      & energy unit arrivals                                                  \\\hline
              $\bm\chi$, $\bm\chi^j$                        & network state of the MU                                               \\\hline
              $\bm\Phi$                                     & control policies of the MU                                            \\\hline
              $c$, $c^j$                                    & task offloading decision by the MU                                    \\\hline
              $e$, $e^j$                                    & energy unit allocation by the MU                                      \\\hline
              $s$, $s^j$                                    & MU-BS association state                                               \\\hline
              $u$                                           & utility function of the MU                                            \\\hline
              $V$                                           & state-value function                                                  \\\hline
              $Q$, $Q_k$                                    & state-action value functions                                          \\\hline
              $\bm\theta^j$, $\bm\theta_k^j$                & DQN parameters                                                        \\\hline
              $\bm\theta_{-}^j$, $\bm\theta_{k, -}^j$       & target DQN parameters                                                 \\\hline
              $\mathcal{M}^j$, $\mathcal{N}^j$              & pools of historical experience                                        \\\hline
              $\widetilde{\mathcal{M}}^j$,
              $\widetilde{\mathcal{N}}^j$                   & mini-batches for DQN training                                         \\
              \hline
        \end{tabular}
    \end{center}
\end{table}

\section{System Descriptions and Assumptions}
\label{sysm}

\begin{figure}[t]
  \centering
  \includegraphics[width=30pc]{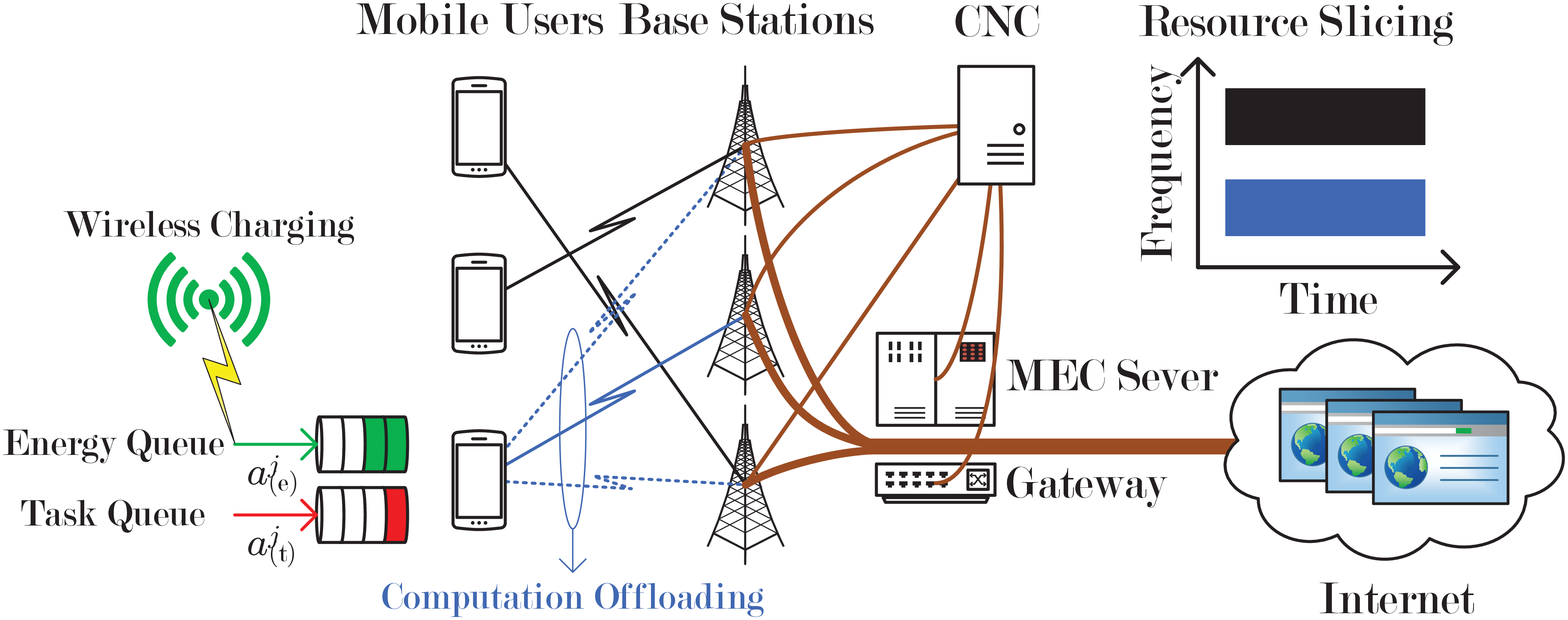}
  \caption{Illustration of mobile-edge computing (MEC) in a virtualized radio access network, where the devices of mobile users are wireless charging enabled, the radio resource is sliced between conventional communication services (the links in black color) and MEC services (the links in blue color), and a centralized network controller (CNC) is responsible for all control plane decisions over the network.}
  \label{systMode}
\end{figure}

As illustrated in Fig. \ref{systMode}, we shall consider in this paper an ultra dense service area covered by a virtualized RAN with a set $\mathcal{B} = \{1, \cdots, B\}$ of BSs.
Both traditional communication services and MEC services are supported over the common physical network infrastructure.
A MEC server is implemented at the network edge, providing rich computing resources for the MUs.
By strategically offloading the generated computation tasks via the BSs to the MEC server for execution, the MUs can expect a significantly improved computation experience.
We assume that the wireless radio resources are divided into traditional communication and MEC slices to guarantee inter-slice isolation.
All control plane operations happening in such a hybrid network are managed by the CNC.
The focus of this work is to optimize computation performance from a perspective of the MUs, while the design of joint traditional communication and MEC resource allocation is left for our next-step investigation.
In a dense networking area, our analysis hereinafter concentrates on a representative MU.
The time horizon is discretized into decision epochs, each of which is of equal duration $\delta$ (in seconds) and is indexed by an integer $j \in \mathds{N}_+$.
Let $W$ (in Hz) denote the frequency bandwidth allocated to the MEC slice, which is shared among the MUs simultaneously accessing the MEC service.

This work assumes that the mobile device of the MU is wireless charging enabled and the received energy can be stored in an energy queue.
%
%
The computation task generated by the MU across the time horizon form an independent and identically distributed sequence of Bernoulli random variables with a common parameter $\lambda_{(\mathrm{t})} \in [0, 1]$.
We denote $a^j_{(\mathrm{t})} \in \{0, 1\}$ as the task arrival indicator, that is, $a^j_{(\mathrm{t})} = 1$ if a computation task is generated from the MU during a decision epoch $j$ and otherwise $a^j_{(\mathrm{t})} = 0$.
Then, $\textsf{Pr}\!\left\{a^j_{(\mathrm{t})} = 1\right\} = 1 - \textsf{Pr}\!\left\{a^j_{(\mathrm{t})} = 0\right\} = \lambda_{(\mathrm{t})}$, where $\textsf{Pr}\{\cdot\}$ denotes the probability of the occurrence of an event.
We represent a computation task by $(\mu, \nu)$ with $\mu$ and $\nu$ being, respectively, the input data size (in bits) and the total number of CPU cycles required to accomplish the task.
A computation task generated at a current decision epoch can be executed starting from the next epoch.
The generated but not processed computation tasks can be queued at the mobile device of the MU.
Based on a first-in first-out principle, a computation task from the task queue can be scheduled for execution either locally on the mobile device or remotely at the MEC server.
More specifically, at the beginning of each decision epoch $j$, the MU makes a joint control action $(c^j, e^j)$, where $c^j \in \{0\} \cup \mathcal{B}$ is the computation offloading decision and $e^j \in \mathds{N}_+$ is the number of allocated energy units\footnote{An energy unit corresponds to an amount of energy, say, $2 \cdot 10^{-3}$ Joules as in numerical experiments.}.
We have $c^j > 0$ if the MU chooses to offload the scheduled computation task to the MEC server via BS $c^j \in \mathcal{B}$ and $c^j = 0$ if the MU decides to execute the computation task locally on its own mobile device.
Note that when $e^j = 0$, the queued tasks will not be executed.

When a computation task is scheduled for processing locally at the mobile device of the MU during a decision epoch $j$, i.e., $c^j = 0$, the allocated CPU-cycle frequency with $e^j > 0$ energy units can be calculated as
\begin{align}\label{cyclFreq}
  f^j = \sqrt{\frac{e^j}{\tau \cdot \nu}},
\end{align}
where $\tau$ is the effective switched capacitance that depends on chip architecture of the mobile device \cite{Burd96}.
Moreover, the CPU-cycle frequency is constrained by $f^j \leq f_{(\mathrm{CPU})}^{(\max)}$.
Then the time needed for local computation task execution is given by
\begin{align}\label{locaDela}
  d^j_{(\mathrm{mobile})} = \frac{\nu}{f^j},
\end{align}
which decreases as the number of allocated energy units increases.
%
%
%
%

We denote $g_b^j$ as the channel gain state between the MU and a BS $b \in \mathcal{B}$ during each decision epoch $j$, which independently picks a value from a finite state space $\mathcal{G}_b$.
The channel state transitions across the time horizon are modelled as a finite-state discrete-time Markov chain.
At the beginning of a decision epoch $j$, if the MU lets the MEC server execute the scheduled computation task on behalf of the mobile device, the input data of the task needs to be offloaded to the chosen BS $c^j \in \mathcal{B}$.
The MU-BS association has to be first established.
If the chosen BS $c^j$ is different from the previously associated one, a handover between the two BSs hence happens.
Denote $s^j \in \mathcal{B}$ as the MU-BS association state at a decision epoch $j$\footnote{We assume that if the MU processes a computation task locally or no task is executed at a decision epoch $j - 1$, then the MU-BS association does not change, namely, $s^j = s^{j - 1}$. In this case, no handover will be triggered.},
\begin{align}
  s^j = b \cdot
  \mathbf{1}_{\left\{\left\{c^{j - 1} = b, b \in \mathcal{B}\right\} \vee \left\{\left\{c^{j - 1} = 0\right\} \wedge\left\{s^{j - 1} = b\right\}\right\}\right\}},
\end{align}
where the symbols $\vee$ and $\wedge$ mean ``logic OR'' and ``logic AND'', respectively, and $\mathbf{1}_{\{\Omega\}}$ is the indicator function that equals 1 if the condition $\Omega$ is satisfied and otherwise 0.
We assume that the energy consumption during the handover procedure is negligible at the mobile device.
In our considered dense networking scenario, the achievable data rate can be written as
\begin{align}\label{dataRate}
  r^j = W \cdot \log_2\!\!\left(1 + \frac{g^j_b \cdot p^j_{(\mathrm{tr})}}{I}\right),
\end{align}
where $I$ is the received average power of interference plus additive background Gaussian noise and
\begin{align}\label{tranPowe}
  p^j_{(\mathrm{tr})} = \frac{e^j}{d^j_{(\mathrm{tr})}},
\end{align}
is the transmit power with
\begin{align}\label{tranTime}
  d^j_{(\mathrm{tr})} = \frac{\mu}{r^j},
\end{align}
being the time of transmitting task input data.
The transmit power is constrained by the maximum transmit power of the mobile device $p_{(\mathrm{tr})}^{(\max)}$ \cite{Kwat16}, i.e.,
\begin{align}\label{tranPoweCons}
  p^j_{(\mathrm{tr})} \leq p_{(\mathrm{tr})}^{(\max)}.
\end{align}

In (\ref{tranTime}) above, we assume that the energy is evenly assigned to the input data bits of the computation task.
In other words, the transmission rate keeps unchanged during the input data transmission.
Lemma 1 ensures that $d^j_{(\mathrm{tr})}$ is the minimum transmission time given the allocated energy units $e^j > 0$.
%
%
%
%
%
%
%
%

\emph{Lemma 1:}
Given the computation offloading decision $c^j \in \mathcal{B}$ and the allocated energy units $e^j > 0$ at a decision epoch $j$, the optimal transmission policy achieving the minimum transmission time is a policy with which the rate of transmitting task input data remains a constant.

\emph{Proof:}
With the decisions of computation offloading $c^j \in \mathcal{B}$ and energy unit allocation $e^j > 0$ at an epoch $j$, the input data transmission is independent of the network dynamics.
Suppose there exists a new transmission policy with which the MU changes its transmission rate from $r^{j, (1)}$ to $r^{j, (2)} \neq r^{j, (1)}$ at a certain point during the task input data transmission.
We denote the time durations corresponding to transmission rates $r^{j, (1)}$ and $r^{j, (2)}$ as $d^{j, (1)}_{(\mathrm{tr})}$ and $d^{j, (2)}_{(\mathrm{tr})}$, respectively.
Taking into account the maximum transmit power of the mobile device (\ref{tranPoweCons}), it is easy to verify that the following two constraints on total energy consumption and total transmitted data bits can be satisfied:
\begin{align}
     \frac{I}{g^j_{c^j}} \!\cdot\! \left(2^{\frac{r^{j, (1)}}{W}} - 1\right) \!\cdot\! d^{j, (1)}_{(\mathrm{tr})}
   + \frac{I}{g^j_{c^j}} \!\cdot\! \left(2^{\frac{r^{j, (2)}}{W}} - 1\right) \!\cdot\! d^{j, (2)}_{(\mathrm{tr})}
 & = \min\!\left\{e^j, \left(d^{j, (1)}_{(\mathrm{tr})} + d^{j, (2)}_{(\mathrm{tr})}\right) \!\cdot\! p_{(\mathrm{tr})}^{(\max)}\right\},   \\
     r^{j, (1)} \cdot d^{j, (1)}_{(\mathrm{tr})} + r^{j, (2)} \cdot d^{j, (2)}_{(\mathrm{tr})}
 & = \mu.
\end{align}
On the other hand, the average transmission rate within a duration of $d^{j, (1)}_{(\mathrm{tr})} + d^{j, (2)}_{(\mathrm{tr})}$ can be written as
\begin{align}
  \bar{r}^j = W \cdot \log_2\!\!\left(1 + \frac{g^j_{c^j}}{I} \cdot
              \frac{\min\!\left\{e^j, \left(d^{j, (1)}_{(\mathrm{tr})} + d^{j, (2)}_{(\mathrm{tr})}\right) \cdot p_{(\mathrm{tr})}^{(\max)}\right\}}
              {d^{j, (1)}_{(\mathrm{tr})} + d^{j, (2)}_{(\mathrm{tr})}}\right).
\end{align}
Consider using $\bar{r}^j$ as the only transmission rate during the whole period of task input data transmission, we construct the following deduction
\begin{align}\label{dedu1}
 &       \bar{r}^j \cdot \left(d^{j, (1)}_{(\mathrm{tr})} + d^{j, (2)}_{(\mathrm{tr})}\right)                                           \nonumber\\
 & =     W \cdot \log_2\!\!\left(1 + \frac{\left(2^{\frac{r^{j, (1)}}{W}} - 1\right) \cdot
         d^{j, (1)}_{(\mathrm{tr})} + \left(2^{\frac{r^{j, (2)}}{W}} - 1\right)
   \cdot d^{j, (2)}_{(\mathrm{tr})}}{d^{j, (1)}_{(\mathrm{tr})} + d^{j, (2)}_{(\mathrm{tr})}}\right) \cdot
         \left(d^{j, (1)}_{(\mathrm{tr})} + d^{j, (2)}_{(\mathrm{tr})}\right)                                                           \nonumber\\
 & \geq  \left(r^{j, (1)} \cdot \frac{d^{j, (1)}_{(\mathrm{tr})}}{d^{j, (1)}_{(\mathrm{tr})} + d^{j, (2)}_{(\mathrm{tr})}} +
         r^{j, (2)} \cdot \frac{d^{j, (2)}_{(\mathrm{tr})}}{d^{j, (1)}_{(\mathrm{tr})} + d^{j, (2)}_{(\mathrm{tr})}}\right) \cdot
         \left(d^{j, (1)}_{(\mathrm{tr})} + d^{j, (2)}_{(\mathrm{tr})}\right)                                                           \nonumber\\
 & =     r^{j, (1)} \cdot d^{j, (1)}_{(\mathrm{tr})} + r^{j, (2)} \cdot d^{j, (2)}_{(\mathrm{tr})} = \mu,
\end{align}
where the inequality originates from the concavity of a logarithmic function.
From (\ref{dedu1}), we can find that with a constant transmission rate, more data can be transmitted within the same transmission period.
Hence, applying a constant rate for the transmission of $\mu$ task input data bits achieves the minimum transmission time, which can be solved according to (\ref{dataRate}) and (\ref{tranTime}).
This completes the proof.
\hfill$\Box$

\emph{Lemma 2:}
Given the computation offloading decision $c^j \in \mathcal{B}$ at a decision epoch $j$, the input data transmission time $d^j_{(\mathrm{tr})}$ is a monotonically decreasing function of the allocated energy units $e^j > 0$.

\emph{Proof:}
By replacing $r^j$ in (\ref{dataRate}) with (\ref{tranTime}), we get
\begin{align}\label{tranSolu}
  \log_2\!\!\left(1 + \frac{g^j_{c^j}}{I} \cdot e^j \cdot \frac{1}{d^j_{(\mathrm{tr})}}\right) = \frac{\mu}{W} \cdot \frac{1}{d^j_{(\mathrm{tr})}}.
\end{align}
Alternatively, we take $\frac{1}{d^j_{(\mathrm{tr})}}$ as the solution of (\ref{tranSolu}), which is an intersection of two lines, namely, $\ell_1(x) = \log_2\left(1 + \frac{g^j_{c^j}}{I} \cdot e^j \cdot x\right)$ and $\ell_2(x) = \frac{\mu}{W} \cdot x$, for $x > 0$.
From the non-negativity and the monotonicity of a logarithmic function and a linear function, it is easy to see that $\frac{1}{d^j_{(\mathrm{tr})}}$ is a monotonically increasing function of $e^j$.
Thus, $d^j_{(\mathrm{tr})}$ is a monotonically decreasing function of $e^j$.
\hfill$\Box$

In addition, we assume in this paper that the battery capacity at the mobile device of the MU is limited and the received energy units across the time horizon take integer values.
Let $q^j_{(\mathrm{e})}$ be the energy queue length of the MU at the beginning of a decision epoch $j$, which evolves according to
\begin{align}\label{enerQueu}
  q^{j + 1}_{(\mathrm{e})} = \min\!\left\{q^j_{(\mathrm{e})} - e^j + a^j_{(\mathrm{e})}, q_{(\mathrm{e})}^{(\max)}\right\},
\end{align}
where $q_{(\mathrm{e})}^{(\max)} \in \mathds{N}_+$ denotes the battery capacity limit and $a^j_{(\mathrm{e})} \in \mathds{N}_+$ is the number of energy units received by the end of decision epoch $j$.

\section{Problem Formulation}
\label{prob}

In this section, we shall first formulate the problem of stochastic computation offloading within the MDP framework and then discuss the optimal solutions.

\subsection{Stochastic Computation Task Offloading}

The experienced delay is the key performance indicator for evaluating the quality of a task computing experience.
The delay of a computation task is defined as the period of time from when the task arrives to the computation task queue to when the task is successfully removed from the task queue.
Thus the experienced delay includes the computation task execution delay and the task queuing delay.
We assume that there is a delay of $\zeta$ seconds for control signalling during the occurrence of one handover.
With a joint control action $(c^j, e^j)$ at a decision epoch $j$, the handover delay can be then given as
\begin{align}\label{handDela}
    h^j = \zeta \cdot \mathbf{1}_{\left\{\left\{c^j \in \mathcal{B}\right\} \wedge \left\{c^j \neq s^j\right\}\right\}}.
\end{align}
According to (\ref{locaDela}), (\ref{tranTime}) and (\ref{handDela}), we obtain the task execution delay as\footnote{In this work, we assume that the BSs are connected to the MEC server via fibre links.
Hence, the round-trip delay between the BSs and the MEC server is negligible.
Further, we neglect the time overhead for the selected BS to send back the computation result due to the fact that the size of a computation outcome is much smaller than the input data of a computation task \cite{Chen16}.}
\begin{align}\label{dela}
  d^j =
  \left\{\!
  \begin{array}{l@{~}l}
     d^j_{(\mathrm{mobile})},                           & \mbox{if } e^j > 0 \mbox{ and } c^j = 0;              \\
     h^j + d^j_{(\mathrm{tr})} + d_{(\mathrm{server})}, & \mbox{if } e^j > 0 \mbox{ and } c^j \in \mathcal{B};  \\
     0,                                                 & \mbox{if } e^j = 0,
  \end{array}
  \right.
\end{align}
where $d_{(\mathrm{server})}$ is time consumed for task execution at the MEC server.
Due to the sufficient available computation resource at the MEC server, we assume that $d_{(\mathrm{server})}$ is a sufficiently small constant.

Notice that if: 1) the MU fails to process a computation task at the mobile device within one decision epoch; or 2) a computation task is scheduled for MEC server execution but the computation result cannot be sent back via the chosen BS within the decision epoch, the task execution fails and the task will remain in the queue until being successfully executed.
The dynamics of the computation task queue at the MU can be hence expressed as
\begin{align}\label{taskQueu}
 q^{j + 1}_{(\mathrm{t})} =
 \min\!\left\{q^j_{(\mathrm{t})} - \mathbf{1}_{\left\{0 < d^j \leq \delta\right\}} + a^j_{(\mathrm{t})}, q_{(\mathrm{t})}^{(\max)}\right\},
\end{align}
where $q^j_{(\mathrm{t})}$ is the number of computation tasks in the queue at the beginning of each decision epoch $j$ and $q_{(\mathrm{t})}^{(\max)} \in \mathds{N}_+$ limits the maximum number of computation tasks that can be queued at the mobile device.
There will be computation task drops once the task queue is full.
We let
\begin{align}\label{taskDrop}
  \eta^j = \max\!\left\{q^j_{(\mathrm{t})} - \mathbf{1}_{\left\{0 < d^j \leq \delta\right\}} + a^j_{(\mathrm{t})} - q_{(\mathrm{t})}^{(\max)}, 0\right\},
\end{align}
define a computation task drop.

If a computation task remains in the queue for a decision epoch, a delay of $\delta$ seconds will be incurred to the task.
We treat the queuing delay during a decision epoch $j$ equivalently as the length of a task queue, that is,
\begin{align}\label{overDela}
   \rho^j = q^j_{(\mathrm{t})} - \mathbf{1}_{\left\{d^j > 0\right\}}.
\end{align}
%
%
%
As previously discussed, if $d^j > \delta$, the execution of a computation task fails.
In this case, the MU receives a penalty, which is defined by
\begin{align}\label{pena}
  \varphi^j = \mathbf{1}_{\left\{d^j > \delta\right\}}.
\end{align}
Moreover, a payment is required for the access to MEC service when the MU decides to offload a computation task for MEC server execution.
The payment is assumed to be proportional to the time consumed for transmitting and processing the task input data.
That is, the payment can be calculated as
\begin{align}\label{paym}
  \phi^j = \pi \cdot \left(\min\!\left\{d^j, \delta\right\} - h^j\right) \cdot \mathbf{1}_{\left\{c^j \in \mathcal{B}\right\}},
\end{align}
where $\pi \in \mathds{R}_+$ is the price paid for the MEC service per unit of time.

The network state of the MU during each decision epoch $j$ can be characterized by $\bm\chi^j = \left(q^j_{(\mathrm{t})}, q^j_{(\mathrm{e})}, s^j, \mathbf{g}^j\right) \in \mathcal{X} \overset{\mathrm{def}}{=} \left\{0, 1, \cdots, q_{(\mathrm{t})}^{(\max)}\right\} \times \left\{0, 1, \cdots, q_{(\mathrm{e})}^{(\max)}\right\} \times \mathcal{B} \times \left\{\times_{b \in \mathcal{B}} \mathcal{G}_b\right\}$, where $\mathbf{g}^j = \left(g^j_b: b \in \mathcal{B}\right)$.
At the beginning of epoch $j$, the MU decides a joint task offloading and energy allocation decision $(c^j, e^j) \in \mathcal{Y} \overset{\mathrm{def}}{=} \{\{0\} \cup \mathcal{B}\} \times \left\{0, 1, \cdots, Q_{(\mathrm{e})}\right\}$\footnote{To keep what follows uniform, we do not exclude the infeasible joint actions.} according to the stationary control policy defined by Definition 1.
In line with the discussions, we define an immediate utility at epoch $j$ to quantify the task computation experience for the MU,
\begin{align}\label{utilFunc}
     u\!\left(\bm\chi^j, (c^j, e^j)\right)
 & = \omega_1 \cdot u^{(1)}\!\left(\min\!\left\{d^j, \delta\right\}\right) + \omega_2 \cdot u^{(2)}(\eta^j) \nonumber\\
 & + \omega_3 \cdot u^{(3)}(\rho^j) + \omega_4 \cdot u^{(4)}(\varphi^j) +  \omega_5 \cdot u^{(5)}(\phi^j),
\end{align}
where the positive monotonically deceasing functions $u^{(1)}(\cdot)$, $u^{(2)}(\cdot)$, $u^{(3)}(\cdot)$, $u^{(4)}(\cdot)$ and $u^{(5)}(\cdot)$ measure the satisfactions of the task execution delay, the computation task drops, the task queuing delay, the penalty of failing to execute a computation task and the payment of accessing the MEC service, and $\omega_1$, $\omega_2$, $\omega_3$, $\omega_4$, $\omega_5 \in \mathds{R}_+$ are the weights that combine different types of function with different units into a universal utility function.
With slight abuse of notations, we rewrite $u^{(1)}(\cdot)$, $u^{(2)}(\cdot)$, $u^{(3)}(\cdot)$, $u^{(4)}(\cdot)$ and $u^{(5)}(\cdot)$ as $u^{(1)}\!\left(\bm\chi^j, (c^j, e^j)\right)$, $u^{(2)}\!\left(\bm\chi^j, (c^j, e^j)\right)$, $u^{(3)}\!\left(\bm\chi^j, (c^j, e^j)\right)$, $u^{(4)}\!\left(\bm\chi^j, (c^j, e^j)\right)$ and $u^{(5)}\!\left(\bm\chi^j, (c^j, e^j)\right)$.

\emph{Definition 1 (Joint Task Offloading and Energy Allocation Control Policy):}
A stationary joint task offloading and energy allocation control policy $\bm\Phi$ is defined as a mapping: $\bm\Phi: \mathcal{X} \rightarrow \mathcal{Y}$.
More specifically, the MU determines a joint control action
$\bm\Phi(\bm\chi^j) = \left(\Phi_{(\mathrm{c})}(\bm\chi^j), \Phi_{(\mathrm{e})}(\bm\chi^j)\right) = (c^j, e^j) \in \mathcal{Y}$
according to $\bm\Phi$ after observing network state $\bm\chi^j \in \mathcal{X}$ at the beginning of each decision epoch $j$,
where $\bm\Phi = \left(\Phi_{(\mathrm{c})}, \Phi_{(\mathrm{e})}\right)$ with $\Phi_{(\mathrm{c})}$ and $\Phi_{(\mathrm{e})}$ being, respectively, the stationary task offloading and energy allocation policies.

Given a stationary control policy $\bm\Phi$, the $\{\bm\chi^j: j \in \mathds{N}_+\}$ is a controlled Markov chain with the following state transition probability
\begin{align}\label{statTran}
\begin{array}{r@{~}c@{~}l}
           \displaystyle\textsf{Pr}\!\left\{\bm\chi^{j + 1} | \bm\chi^j, \bm\Phi\!\left(\bm\chi^j\right)\right\}
 & =     & \displaystyle\textsf{Pr}\!\left\{q^{j + 1}_{(\mathrm{t})} | q^j_{(\mathrm{t})}, \bm\Phi\!\left(\bm\chi^j\right)\right\} \cdot
           \textsf{Pr}\!\left\{q^{j + 1}_{(\mathrm{e})} | q^j_{(\mathrm{e})}, \bm\Phi\!\left(\bm\chi^j\right)\right\} \\
 & \cdot & \displaystyle\textsf{Pr}\!\left\{s^{j + 1} | s^j, \bm\Phi\!\left(\bm\chi^j\right)\right\} \cdot
           \prod\limits_{b \in \mathcal{B}} \textsf{Pr}\!\left\{g^{j + 1}_b | g^j_b\right\}.
\end{array}
\end{align}
Taking expectation with respect to the per-epoch utilities $\left\{u\!\left(\bm\chi^j, \bm\Phi\!\left(\bm\chi^j\right)\right): j \in \mathds{N}_+\right\}$ over the sequence of network states $\left\{\bm\chi^j: j \in \mathds{N}_+\right\}$, the expected long-term utility of the MU conditioned on an initial network state $\bm\chi^1$ can be expressed as
\begin{align}\label{expeCost}
   V(\bm\chi, \bm\Phi) =
   \textsf{E}_{\bm\Phi}\!\!\left[(1 - \gamma) \cdot \sum_{j = 1}^\infty (\gamma)^{j - 1} \cdot u\!\left(\bm\chi^j, \bm\Phi\left(\bm\chi^j\right)\right) | \bm\chi^1 = \bm\chi\right],
\end{align}
where $\bm\chi = \left(q_{(\mathrm{t})}, q_{(\mathrm{e})}, s, \mathbf{g}\right) \in \mathcal{X}$, $\mathbf{g} = (g_b: b \in \mathcal{B})$, $\gamma \in [0, 1)$ is the discount factor, and $(\gamma)^{j-1}$ denotes the discount factor to the $(j-1)$-th power.
$V(\bm\chi, \bm\Phi)$ is also named as the state-value function for the MU in the state $\bm\chi$ under policy $\bm\Phi$.

\emph{Problem 1:}
The MU aims to design an optimal stationary control policy $\bm\Phi^* = \left(\Phi_{(\mathrm{c})}^*, \Phi_{(\mathrm{e})}^*\right)$ that maximizes the expected long-term utility performance, $V(\bm\chi, \bm\Phi)$, for any given initial network state $\bm\chi$, which can be formally formulated as in the following
\begin{align}\label{optiPoli}
  \bm\Phi^* = \underset{\bm\Phi}{\arg\max}~V(\bm\chi, \bm\Phi), \forall \bm\chi \in \mathcal{X}.
\end{align}
$V(\bm\chi) = V\!\left(\bm\chi, \bm\Phi^*\right)$ is defined as the optimal state-value function, $\forall \bm\chi \in \mathcal{X}$.

\emph{Remark 1:}
The formulated problem of stochastic computation offloading optimization as in Problem 1 is in general a single-agent infinite-horizon MDP with the discounted utility criterion.
Nevertheless, (\ref{expeCost}) can also be used to approximate the expected infinite-horizon undiscounted utility \cite{Adel08}
\begin{align}
   U(\bm\chi, \bm\Phi) =
   \textsf{E}_{\bm\Phi}\!\!\left[\lim_{J \rightarrow \infty} \frac{1}{J} \cdot \sum_{j = 1}^J u\!\left(\bm\chi^j, \bm\Phi\left(\bm\chi^j\right)\right) | \bm\chi^1 = \bm\chi\right].
\end{align}
when $\gamma$ approaches $1$.

\subsection{Learning Optimal Solution to Problem 1}

The stationary control policy achieving the optimal state-value function can be obtained by solving the following Bellman's optimality equation \cite{Rich98}: $\forall \bm\chi \in \mathcal{X}$,
\begin{align}\label{BellEqua}
    V(\bm\chi)
  = \max_{(c, e)}\left\{(1 - \gamma) \cdot u(\bm\chi, (c, e))
  + \gamma \cdot \sum_{\bm\chi'} \textsf{Pr}\!\left\{\bm\chi' | \bm\chi, (c, e)\right\} \cdot V(\bm\chi')\right\},
\end{align}
where $u(\bm\chi, (c, e))$ is the achieved utility when a joint control action $(c, e) \in \mathcal{Y}$ is performed under network state $\bm\chi$ and $\bm\chi' = \left(q_{(\mathrm{t})}', q_{(\mathrm{e})}', s', \mathbf{g}'\right) \in \mathcal{X}$ is the subsequent network state with $\mathbf{g}' = \left(g_b': b \in \mathcal{B}\right)$.

\textit{Remark 2:}
The traditional solutions to (\ref{BellEqua}) are based on the value iteration or the policy iteration \cite{Rich98}, which need \emph{complete knowledge} of the computation task arrival, the received energy unit and the channel state transition statistics.

One attractiveness of the off-policy $Q$-learning is that it assumes no a priori knowledge of the network state transition statistics \cite{Rich98}.
We define the right-hand side of (\ref{BellEqua}) by
\begin{align}\label{stat_acti_q1}
   Q(\bm\chi, (c, e))
 = (1 - \gamma) \cdot u(\bm\chi, (c, e))
 + \gamma \cdot \sum_{\bm\chi'} \textsf{Pr}\{\bm\chi' | \bm\chi, (c, e)\} \cdot V(\bm\chi'),
\end{align}
$\forall \bm\chi \in \mathcal{X}$.
The optimal state-value function $V(\bm\chi)$ can be hence directly obtained from
\begin{align}\label{stat_acti_q2}
    V(\bm\chi) = \max_{(c, e)} Q(\bm\chi, (c, e)).
\end{align}
By substituting (\ref{stat_acti_q2}) into (\ref{stat_acti_q1}), we get
\begin{align}\label{stat_acti_q3}
  Q(\bm\chi, (c, e)) =
  (1 - \gamma) \cdot u(\bm\chi, (c, e)) +
  \gamma \cdot \sum_{\bm\chi'} \textsf{Pr}\{\bm\chi' | \bm\chi, (c, e)\} \cdot \max_{(c', e')} Q(\bm\chi', (c', e')),
\end{align}
where we denote $(c', e') \in \mathcal{Y}$ as a joint control action performed under the network state $\bm\chi'$.
In practice, the computation task arrival as well as the number of energy units that can be received by the end of a decision epoch are unavailable beforehand.
Using standard $Q$-learning, the MU tries to learn $Q(\bm\chi, (c, e))$ in a recursive way based on the observation of network state $\bm\chi = \bm\chi^j$ at a current decision epoch $j$, the performed joint action $(c, e) = (c^j, e^j)$, the achieved utility $u(\bm\chi, (c, e))$ and the resulting network state $\bm\chi' = \bm\chi^{j + 1}$ at the next epoch $j + 1$.
The updating rule is
\begin{align}\label{QLearRule}
     Q^{j + 1}(\bm\chi, (c, e))
 & = Q^j(\bm\chi, (c, e))                                                                                                                       \nonumber\\
 & + \alpha^j \left((1 - \gamma) \cdot u(\bm\chi, (c, e)) + \gamma \cdot \max_{(c', e')} Q^j(\bm\chi', (c', e')) - Q^j(\bm\chi, (c, e))\right),
\end{align}
where $\alpha^j \in [0, 1)$ is a time-varying learning rate.
It has been proven that if 1) the network state transition probability under the optimal stationary control policy is stationary, 2) $\sum_{j = 1}^\infty \alpha^j$ is infinite and $\sum_{j = 1}^\infty (\alpha^j)^2$ is finite, and 3) all state-action pairs are visited infinitely often, the $Q$-learning process converges and eventually finds the optimal control policy \cite{Watk12}.
The last condition can be satisfied if the probability of choosing any action in any network state is non-zero (i.e., \emph{exploration}).
Meanwhile, the MU has to exploit the most recent knowledge of $Q$-function in order to perform well (i.e., \emph{exploitation}).
A classical way to balance the trade-off between \emph{exploration} and \emph{exploitation} is the $\epsilon$-greedy strategy \cite{Rich98}.

\textit{Remark 3:}
From (\ref{QLearRule}), we can find that the standard $Q$-learning rule suffers from \emph{poor scalability}.
Due to the tabular nature in representing $Q$-function values, $Q$-learning is not readily applicable to high-dimensional scenarios with extremely huge network state and/or action spaces, where the learning process is extremely slow.
In our considered system model, the sizes of the network state space $\mathcal{X}$ and the action space $\mathcal{Y}$ can be calculated as $X = \left(1 + q_{(\mathrm{t})}^{(\max)}\right) \cdot \left(1 + q_{(\mathrm{e})}^{(\max)}\right) \cdot B \cdot \prod_{b \in \mathcal{B}} |\mathcal{G}_b|$ and $Y = (1 + B) \cdot \left(1 + q_{(\mathrm{e})}^{(\max)}\right)$, respectively, where $|\mathcal{G}|$ means the cardinality of the set $\mathcal{G}$.
It can be observed that $X$ grows exponentially as the number $B$ of BSs increases.
Suppose there is a MEC system with $6$ BSs and for each BS, the channel gain is quantized into $6$ states (as assumed in our experiment setups).
If we set $q_{(\mathrm{t})}^{(\max)} = q_{(\mathrm{e})}^{(\max)} = 4$, the MU has to update in total $X \cdot Y = 2.44944 \cdot 10^8$ $Q$-function values during the learning process, which is impossible for the $Q$-learning process to converge within limited number of decision epoches.

The next section thereby focuses on developing practically feasible and computationally efficient algorithms to approach the optimal control policy.

\section{Approaching the Optimal Policy}
\label{solu}

In this section, we proceed to approach the optimal control policy by developing practically feasible algorithms based on recent advances in deep reinforcement learning and a linear $Q$-function decomposition technique.

\subsection{Deep Reinforcement Learning Algorithm}
\label{darling}

\begin{figure}[t]
  \centering
  \includegraphics[width=26pc]{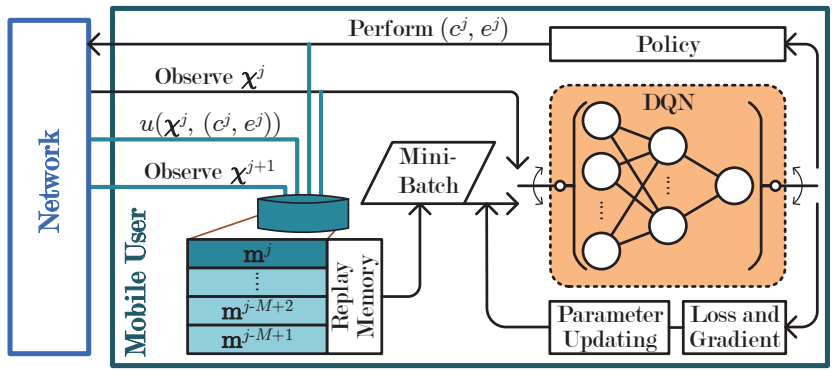}
  \caption{Double deep $Q$-network (DQN) based reinforcement learning (DARLING) for stochastic computation offloading in a mobile-edge computing system.}
  \label{deepLear}
\end{figure}

Inspired by the success of modelling an optimal state-action $Q$-function with a deep neural network \cite{Mnih15}, we adopt a double DQN to address the massive network state space $\mathcal{X}$ \cite{Hass16}.
Specifically, the $Q$-function expressed as in (\ref{stat_acti_q1}) is approximated by $Q(\bm\chi, (c, e)) \approx Q(\bm\chi, (c, e) ; \bm\theta)$, where $(\bm\chi, (c, e)) \in \mathcal{X} \times \mathcal{Y}$ and $\bm\theta$ denotes a vector of parameters associated with the DQN.
%
%
The DQN-based reinforcement learning (DARLING) for stochastic computation offloading in our considered MEC system is illustrated in Fig. \ref{deepLear}, during which instead of finding the optimal $Q$-function, the DQN parameters can be learned iteratively.

The mobile device is assumed to be equipped with a replay memory of a finite size $M$ to store the experience $\mathbf{m}^j = \left(\bm\chi^j, (c^j, e^j), u(\bm\chi^j, (c^j, e^j)), \bm\chi^{j + 1}\right)$ that is happened at the transition of two consecutive decision epoches $j$ and $j + 1$ during the learning process of DARLING, where $\bm\chi^j$, $\bm\chi^{j + 1} \in \mathcal{X}$ and $(c^j, e^j) \in \mathcal{Y}$.
The experience pool can be represented as $\mathcal{M}^j = \left\{\mathbf{m}^{j - M + 1}, \cdots, \mathbf{m}^j\right\}$.
The MU maintains a DQN and a target DQN, namely, $Q\!\left(\bm\chi, (c, e) ; \bm\theta^j\right)$ and $Q\!\left(\bm\chi, (c, e) ; \bm\theta_{-}^j\right)$, with parameters $\bm\theta^j$ at a current decision epoch $j$ and $\bm\theta_{-}^j$ at some previous epoch before decision epoch $j$, $\forall (\bm\chi, (c, e)) \in \mathcal{X} \times \mathcal{Y}$.
According to the experience replay technique \cite{Lin92}, the MU then randomly samples a mini-batch $\widetilde{\mathcal{M}}^j \subseteq \mathcal{M}^j$ from the pool $\mathcal{M}^j$ of historical experiences at each decision epoch $j$ to online train the DQN.
That is, the parameters $\bm\theta^j$ are updated in the direction of minimizing the loss function, which is defined by
\begin{align}\label{lossFunc}
 & L_{(\mathrm{DARLING})}\!\left(\bm\theta^j\right) =
        &       &                                                                                                                                       \nonumber\\
 & \textsf{E}_{\left(\bm\chi, (c, e), u(\bm\chi, (c, e)), \bm\chi'\right) \in \widetilde{\mathcal{M}}^j}\!\! \Bigg[\! \bigg(\!
        &       & \!\!\!\!\! (1 - \gamma) \cdot u(\bm\chi, (c, e)) + \gamma \cdot
                  Q\!\left(\bm\chi', \underset{(c', e')}{\arg\max}~\mathds{Q}\!\left(\bm\chi', (c', e'); \bm\theta^j\right); \bm\theta_{-}^j\right) -   \nonumber\\
 &      &       & \!\!\!\!\! Q\!\left(\bm\chi, (c, e); \bm\theta^j\right)\!\!\bigg)^2\Bigg],
\end{align}
where $(c', e') \in \mathcal{Y}$.
The loss function $L_{(\mathrm{DARLING})}(\bm\theta^j)$ is a mean-squared measure of the Bellman equation error at a decision epoch $j$ (i.e., the last term of (\ref{QLearRule})) by replacing $Q^j(\bm\chi, (c, e))$ and its corresponding target $(1 - \gamma) \cdot u(\bm\chi, (c, e)) + \gamma \cdot \max_{(c', e')} Q^j(\bm\chi', (c', e'))$ with $Q(\bm\chi, (c, e); \bm\theta^j)$ and $(1 - \gamma) \cdot u(\bm\chi, (c, e)) + \gamma \cdot Q\!\left(\bm\chi', \arg\max_{(c', e')} Q(\bm\chi', (c', e'); \bm\theta^j); \bm\theta_{-}^j\right)$ \cite{Hass16}, respectively.
By differentiating the loss function $L_{(\mathrm{DARLING})}(\bm\theta^j)$ with respect to the DQN parameters $\bm\theta^j$, we obtain the gradient as
\begin{align}\label{grad}
 & \nabla_{\bm\theta^j} L_{(\mathrm{DARLING})}\!\left(\bm\theta^j\right) =
        &       &                                                                                                                               \nonumber\\
 & \textsf{E}_{\left(\bm\chi, (c, e), u(\bm\chi, (c, e)), \bm\chi'\right)\in \widetilde{\mathcal{M}}^j}\!\! \bigg[\! \bigg(\!
        &       & \!\!\!\!\! (1 - \gamma) \cdot u(\bm\chi, (c, e)) + \gamma \cdot
                  Q\!\left(\bm\chi', \underset{(c', e')}{\arg\max}~Q\!\left(\bm\chi', (c', e'); \bm\theta^j\right); \bm\theta_{-}^j\right) -    \nonumber\\
 &      &       & \!\!\!\!\! Q\!\left(\bm\chi, (c, e); \bm\theta^j\right)\!\!\bigg) \cdot
                  \nabla_{\bm\theta^j} Q\!\left(\bm\chi, (c, e); \bm\theta^j\right)\bigg].
\end{align}
Algorithm \ref{algo} summarizes the implementation of the online DARLING algorithm by the MU for stochastic computation offloading in our considered MEC system.
\begin{algorithm}[t]
    \caption{Online DARLING Algorithm for Stochastic Computation Task Offloading in A MEC System}
    \label{algo}
    \begin{algorithmic}[1]
        \STATE \textbf{initialize} the replay memory $\mathcal{M}^j$ with a finite size of $M \in \mathds{N}_+$, the mini-batch $\widetilde{\mathcal{M}}^j$ with a size of $\tilde{M} < M$ for experience replay, a DQN and a target DQN with two sets $\bm\theta^j$ and $\bm\theta_{-}^j$ of random parameters, for $j = 1$.

        \REPEAT
            \STATE At the beginning of decision epoch $j$, the MU observes the network state $\bm\chi^j \in \mathcal{X}$, which is taken as an input to the DQN with parameters $\bm\theta^j$, and then selects a joint control action $(c^j, e^j) \in \mathcal{Y}$ randomly with probability $\epsilon$ or $(c^j, e^j) = \arg\max_{(c, e) \in \mathcal{Y}} Q(\bm\chi^j, (c, e); \bm\theta^j)$ with probability $1 - \epsilon$.

            \STATE After performing the selected joint control action $(c^j, e^j)$, the MU realizes an immediate utility $u(\bm\chi^j, (c^j, e^j))$ and observes the new network state $\bm\chi^{j + 1} \in \mathcal{X}$ at the next decision epoch $j + 1$.

            \STATE The MU updates the replay memory $\mathcal{M}^j$ at the mobile device with the most recent transition $\mathbf{m}^j = \left(\bm\chi^j, (c^j, e^j), u(\bm\chi^j, (c^j, e^j)), \bm\chi^{j + 1}\right)$.

            \STATE With a randomly sampled mini-batch of transitions $\widetilde{\mathcal{M}}^j \subseteq \mathcal{M}^j$ from the replay memory, the MU updates the DQN parameters $\bm\theta^j$ with the gradient given by (\ref{grad}).

            \STATE The MU regularly reset the target DQN parameters with $\bm\theta_{-}^{j + 1} = \bm\theta^j$, and otherwise $\bm\theta_{-}^{j + 1} = \bm\theta_{-}^j$.

            \STATE The decision epoch index is updated by $j \leftarrow j + 1$.
        \UNTIL{A predefined stopping condition is satisfied.}
    \end{algorithmic}
\end{algorithm}

\subsection{Linear $Q$-Function Decomposition based Deep Reinforcement Learning}
\label{Line}

\subsubsection{Linear $Q$-Function Decomposition}

It can be found that the utility function in (\ref{utilFunc}) is of an additive structure, which motivates us to linearly decompose the state-action $Q$-function, namely, $Q(\bm\chi, (c, e))$, $\forall (\bm\chi, (c, e)) \in \mathcal{X} \times \mathcal{Y}$, based on the pattern $\mathcal{K} = \{1, \cdots, K\}$ of classifying the satisfactions regarding the task execution delay, the computation task drops, the task queuing delay, the penalty of failing to process a computation task and the payment of using the MEC service.
For example, we can divide the utility into four satisfaction categories, namely, $u\!\left(\bm\chi, (c, e)\right) = u_1\!\left(\bm\chi, (c, e)\right) + u_2\!\left(\bm\chi, (c, e)\right) + u_3\!\left(\bm\chi, (c, e)\right) + u_4\!\left(\bm\chi, (c, e)\right)$ with $u_1\!\left(\bm\chi, (c, e)\right) = \omega_1 \cdot u^{(1)}\!\left(\bm\chi, (c, e)\right) + \omega_3 \cdot u^{(3)}\!\left(\bm\chi, (c, e)\right)$, $u_2\!\left(\bm\chi, (c, e)\right) = \omega_2 \cdot u^{(2)}\!\left(\bm\chi, (c, e)\right)$,  $u_3\!\left(\bm\chi, (c, e)\right) = \omega_4 \cdot u^{(4)}\!\left(\bm\chi, (c, e)\right)$ and $u_4\!\left(\bm\chi, (c, e)\right) = \omega_5 \cdot u^{(5)}\!\left(\bm\chi, (c, e)\right)$, then $\mathcal{K} = \{1, 2, 3, 4\}$ forms a classification pattern.
In our considered stochastic computation offloading scenario, it's easy to see that $K \leq 5$.
Mathematically, $Q(\bm\chi, (c, e))$ is decomposed into \cite{Russ03}
\begin{align}\label{Deco}
   Q(\bm\chi, (c, e)) = \sum_{k \in \mathcal{K}} Q_k(\bm\chi, (c, e)),
\end{align}
where the MU deploys a ``virtual'' agent $k \in \mathcal{K}$ to learn the optimal per-agent state-action $Q$-function $Q_k(\bm\chi, (c, e))$ that satisfies
\begin{align}\label{stat_acti_q3_deco}
   Q_k(\bm\chi, (c, e)) = (1 - \gamma) \cdot u_k(\bm\chi, (c, e)) +
   \gamma \cdot \sum_{\bm\chi'} \textsf{Pr}\!\left\{\bm\chi' | \bm\chi, (c, e)\right\} \cdot Q_k\!\left(\bm\chi', \bm\Phi^*(\bm\chi')\right),
\end{align}
with $u_k(\bm\chi, (c, e))$ being the immediate utility related to a satisfaction category $k$.
We emphasize that the optimal joint control action in (\ref{stat_acti_q3_deco}) of an agent $k$ across the time horizon should reflect the optimal control policy implemented by the MU.
In other words, the MU makes an optimal joint control action decision $\bm\Phi^*(\bm\chi)$ under a network state $\bm\chi$
\begin{align}\label{optiActi}
  \bm\Phi^*(\bm\chi) = \underset{(c, e)}{\arg\max} \sum_{k \in \mathcal{K}} Q_k(\bm\chi, (c, e)),
\end{align}
to maximize the aggregated $Q$-function values from all the agents.
We will see in Theorem 1 that the linear $Q$-function decomposition technique achieves the optimal solution to Problem 1.

\emph{Theorem 1:}
The linear $Q$-function decomposition approach as in (\ref{Deco}) asserts the optimal computation offloading performance.

\emph{Proof:}
For the state-action $Q$-function of a joint action $(c, e) \in \mathcal{Y}$ in a network state $\bm\chi \in \mathcal{X}$ as in (\ref{stat_acti_q1}), we have
\begin{align}
     Q(\bm\chi, (c, e))
 & = \textsf{E}_{\bm\Phi^*}\!\!\left[(1 - \gamma) \cdot \sum_{j = 1}^\infty (\gamma)^{j - 1} \cdot
     u\!\left(\bm\chi^j, \left(c^j, e^j\right)\right) | \bm\chi^1 = \bm\chi, \left(c^1, e^1\right) = (c, e)\right]              \nonumber\\
 & = \textsf{E}_{\bm\Phi^*}\!\!\left[(1 - \gamma) \cdot \sum_{j = 1}^\infty (\gamma)^{j - 1} \cdot
     \sum_{k \in \mathcal{K}} u_k\!\left(\bm\chi^j, \left(c^j, e^j\right)\right) |
     \bm\chi^1 = \bm\chi, \left(c^1, e^1\right) = (c, e)\right]                                                                 \nonumber\\
 & = \sum_{k \in \mathcal{K}} \textsf{E}_{\bm\Phi^*}\!\!\left[(1 - \gamma) \cdot \sum_{j = 1}^\infty (\gamma)^{j - 1} \cdot
     u_k\!\left(\bm\chi^j, \left(c^j, e^j\right)\right) |
     \bm\chi^1 = \bm\chi, \left(c^1, e^1\right) = (c, e)\right]                                                                 \nonumber\\
 & = \sum_{k \in \mathcal{K}} Q_k\!\left(\bm\chi, (c, e)\right),
\end{align}
which completes the proof.
\hfill$\Box$

\emph{Remark 4:}
An apparent advantage of the linear $Q$-function decomposition technique is that it potentially simplifies the problem solving.
Back to the example above, agent 2 learns the optimal expected long-term satisfaction measuring the computation task drops across the time horizon.
It's obvious that a computation task drop $\eta^j$ at a decision epoch $j$ depends only on the task queue state $q_{(\mathrm{t})}^j$ (rather than the network state $\bm\chi^j \in \mathcal{X}$) and the performed joint control action $(c^j, e^j)$ by the MU.

Recall that the joint control action of each agent should be in accordance with the optimal control policy of the MU, the $Q$-learning rule, which involves off-policy updates, is obviously not applicable to finding the optimal per-agent state-action $Q$-functions.
The state-action-reward-state-action (SARSA) algorithm \cite{Rumm94, Rich98}, which applies on-policy updates, fosters a promising alternative, namely, a SARSA-based reinforcement learning (SARL).
Having the observations of the network state $\bm\chi = \bm\chi^j$, the performed joint control action $(c, e) = (c^j, e^j)$ by the MU, the realized per-agent utilities $\left(u_k\!\left(\bm\chi, \left(c, e\right)\right): k \in \mathcal{K}\right)$ at a current decision epoch $j$ and the resulting network state $\bm\chi' = \bm\chi^{j + 1}$, the joint control action $(c', e') = (c^{j + 1}, e^{j + 1})$ selected by the MU at the next epoch $j + 1$, each agent $k \in \mathcal{K}$ updates the state-action $Q$-function on the fly,
\begin{align}\label{perPattUpda}
      Q_k^{j + 1}(\bm\chi, (c, e))
  & = Q_k^j(\bm\chi, (c, e))                                                                                                                    \nonumber\\
  & + \alpha^j \cdot \left((1 - \gamma) \cdot u_k(\bm\chi, (c, e)) + \gamma \cdot Q_k^j(\bm\chi', (c', e')) - Q_k^j(\bm\chi, (c, e))\right),
\end{align}
where different from off-policy $Q$-learning, $(c', e')$ is a actually performed joint control action in the subsequent network state, rather than the hypothetical one that maximizes the per-agent state-action $Q$-function.
Theorem 2 ensures that the SARL algorithm converges.

\emph{Theorem 2:}
The sequence $\left\{\left(Q_k^j(\bm\chi, (c, e)): \bm\chi \in \mathcal{X}, (c, e) \in \mathcal{Y} \mbox{~and~} k \in \mathcal{K}\right): j \in \mathbb{N}_+\right\}$ by SARL converges to the optimal per-agent state-action $Q$-functions $Q_k(\bm\chi, (c, e))$, $\forall \bm\chi \in \mathcal{X}$, $\forall (c, e) \in \mathcal{Y}$ and $\forall k \in \mathcal{K}$ if and only if the state-action pairs $(\bm\chi, (c, e)) \in \mathcal{X} \times \mathcal{Y}$ are visited infinitely often and the learning rate $\alpha^j$ satisfies: $\sum_{j = 1}^\infty \alpha^j = \infty$ and $\sum_{j = 1}^\infty \left(\alpha^j\right)^2 < \infty$.

\emph{Proof:}
Since the per-agent state-action $Q$-functions are learned simultaneously, we consider the monolithic updates during the learning process of the SARL algorithm, namely, the updating rule in (\ref{perPattUpda}) can be then encapsulated as
\begin{align}\label{SARSA}
   & \sum_{k \in \mathcal{K}} Q_k^{j + 1}(\bm\chi, (c, e)) =                                                        \\
   & \left(1 - \alpha^j\right) \cdot \sum_{k \in \mathcal{K}} Q_k^j(\bm\chi, (c, e)) +
     \alpha^j \cdot \left((1 - \gamma) \cdot \sum_{k \in \mathcal{K}} u_k(\bm\chi, (c, e)) + \gamma \cdot
     \sum_{k \in \mathcal{K}} Q_k^j(\bm\chi', (c', e'))\right),                                                     \nonumber
\end{align}
where $(\bm\chi, (c, e))$, $(\bm\chi', (c', e')) \in \mathcal{X} \times \mathcal{Y}$.
We rewrite (\ref{SARSA}) as
\begin{align}\label{SARSAv2}
     \sum_{k \in \mathcal{K}} Q_k^{j + 1}(\bm\chi, (c, e)) - \sum_{k \in \mathcal{K}} Q_k(\bm\chi, (c, e))
  & = \left(1 - \alpha^j\right) \cdot \left(\sum_{k \in \mathcal{K}} Q_k^j(\bm\chi, (c, e)) -
      \sum_{k \in \mathcal{K}} Q_k(\bm\chi, (c, e))\right)                                                      \nonumber\\
  & + \alpha^j \cdot \Upsilon^j(\bm\chi, (c, e)),
\end{align}
where
\begin{align}
     \Upsilon^j(\bm\chi, (c, e))
 & = (1 - \gamma) \cdot \sum_{k \in \mathcal{K}} u_k(\bm\chi, (c, e)) + \gamma \cdot
     \max_{(c'', e'')} \sum_{k \in \mathcal{K}} Q_k^j(\bm\chi', (c'', e''))         \\
 & - \sum_{k \in \mathcal{K}} Q_k(\bm\chi, (c, e))) + \gamma \cdot
     \left(\sum_{k \in \mathcal{K}} Q_k^j(\bm\chi', (c', e')) -
     \max_{(c'', e'')} \sum_{k \in \mathcal{K}} Q_k^j(\bm\chi', (c'', e''))\right). \nonumber
\end{align}
Denote $\mathcal{O}^j = \sigma\!\big(\!\{(\bm\chi^z, (c^z, e^z), (u_k(\bm\chi^z, (c^z, e^z))\!: k \in \mathcal{K}))\!: z \leq j\}, \big\{\!Q_k^j(\bm\chi, (c, e)): \forall (\bm\chi, (c, e)) \in$ $\mathcal{X} \times \mathcal{Y}, \forall k \in \mathcal{K}\big\}\big)$ as the learning history for the first $j$ decision epochs.
The per-agent state-action $Q$-functions are $\mathcal{O}^j$-measurable, thus both $(\sum_{k \in \mathcal{K}} Q_k^{j + 1}(\bm\chi, (c, e)) - \sum_{k \in \mathcal{K}} Q_k(\bm\chi, (c, e)))$ and $\Upsilon^j\!\left(\bm\chi, (c, e)\right)$ are $\mathcal{O}^j$-measurable.
We then arrive at
\begin{align}
      & \left\|\textsf{E}\!\left[\Upsilon^j\!\left(\bm\chi, (c, e)\right) | \mathcal{O}^j\right]\right\|_\infty                             \nonumber\\
 \leq & \left\|\textsf{E}\!\left[(1 - \gamma) \cdot \sum_{k \in \mathcal{K}} u_k(\bm\chi, (c, e)) +
                                 \gamma \cdot \max_{(c'', e'')} \sum_{k \in \mathcal{K}} Q_k^j(\bm\chi', (c'', e'')) -
                                 \sum_{k \in \mathcal{K}} Q_k(\bm\chi, (c, e)) | \mathcal{O}^j\right]\right\|_\infty +                      \nonumber\\
      & \left\|\textsf{E}\!\left[\gamma \cdot \left(\sum_{k \in \mathcal{K}} Q_k^j(\bm\chi', (c', e')) -
                                 \max_{(c'', e'')} \sum_{k \in \mathcal{K}} Q_k^j(\bm\chi', (c'', e''))\right) |
                                 \mathcal{O}^j\right]\right\|_\infty                                                                        \nonumber\\
 \overset{\mathrm{(a)}}{\leq}
      & \gamma \cdot \left\|\sum_{k \in \mathcal{K}} Q_k^j(\bm\chi, (c, e)) -
                            \sum_{k \in \mathcal{K}} Q_k(\bm\chi, (c, e))\right\|_\infty +                                                  \nonumber\\
      & \left\|\textsf{E}\!\left[\gamma \cdot \left(\sum_{k \in \mathcal{K}} Q_k^j(\bm\chi', (c', e')) -
                                 \max_{(c'', e'')} \sum_{k \in \mathcal{K}} Q_k^j(\bm\chi', (c'', e''))\right) |
                                 \mathcal{O}^j\right]\right\|_\infty,
\end{align}
where $\mathrm{(a)}$ is due to the convergence property of a standard $Q$-learning \cite{Watk12}.
We are left with verifying that $\left\|\textsf{E}\!\left[\gamma \cdot \left(\sum_{k \in \mathcal{K}} Q_k^j(\bm\chi', (c', e')) - \max_{(c'', e'')} \sum_{k \in \mathcal{K}} Q_k^j(\bm\chi', (c'', e''))\right) | \mathcal{O}^j\right]\right\|_\infty$ converges to zero with probability 1, which in our considered scenario follows from: i) an $\epsilon$-greedy strategy is applied in SARL for choosing joint control actions; ii) the per-agent state-action $Q$-functions are upper bounded; and iii) both the network state and the joint control action spaces are finite.
All conditions in \cite[Lemma 1]{Sing00} are satisfied.
Therefore, the convergence of the SARL learning process is ensured.
\hfill$\Box$

\emph{Remark 5:}
Implementing the derived SARL algorithm, the size of $Q$-function faced by each agent remains the same as the standard $Q$-learning algorithm.
From Remark 4, the linear $Q$-function decomposition technique provides the possibility of simplifying the solving of the stochastic computation offloading problem through introducing multiple ``virtual'' agents.
Each agent learns the respective expected long-term satisfaction by exploiting the key network state information and is hence enabled to use a simpler DQN to approximate the optimal state-action $Q$-function.

\subsubsection{Deep SARSA Reinforcement Learning (Deep-SARL)}

\begin{figure}[t]
  \centering
  \includegraphics[width=26pc]{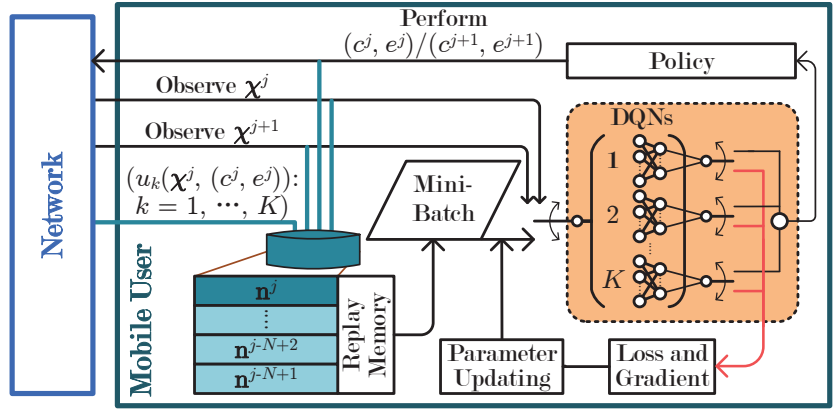}
  \caption{Deep SARSA reinforcement learning (Deep-SARL) based stochastic computation offloading in a mobile-edge computing system.}
  \label{deep_sarl}
\end{figure}

Applying the linear $Q$-function decomposition technique, the DQN $Q(\bm\chi, (c, e); \bm\theta)$, which approximates the optimal state-action $Q$-function $Q(\bm\chi, (c, e))$, $\forall (\bm\chi, (c, e)) \in \mathcal{X} \times \mathcal{Y}$, can be reexpressed as,
\begin{align}\label{Deco_deep}
   Q(\bm\chi, (c, e); \bm\theta) = \sum_{k \in \mathcal{K}} Q_k(\bm\chi, (c, e); \bm\theta_k),
\end{align}
where $\bm\theta = (\bm\theta_k: k \in \mathcal{K})$ is a collection of parameters associated with the DQNs of all agents and $Q_k(\bm\chi, (c, e); \bm\theta_k)$ (for $k \in \mathcal{K}$) is the per-agent DQN.
Accordingly, we derive a novel deep SARSA reinforcement learning (Deep-SARL) based stochastic computation offloading algorithm, as depicted in Fig. \ref{deep_sarl}, where different from the DARLING algorithm, the parameters $\bm\theta$ are learned locally at the agents in an online iterative way.

Implementing our proposed Deep-SARL algorithm, at each decision epoch $j$, the MU updates the experience pool $\mathcal{N}^j$ with the most recent $N$ experiences $\left\{\mathbf{n}^{j - N + 1}, \cdots, \mathbf{n}^j\right\}$ with each experience $\mathbf{n}^j = \left(\bm\chi^j, (c^j, e^j), \left(u_k(\bm\chi^j, (c^j, e^j)): k \in \mathcal{K}\right), \bm\chi^{j + 1}, (c^{j + 1}, e^{j + 1})\right)$.
To train the DQN parameters, the MU first randomly samples a mini-batch $\widetilde{\mathcal{N}}^j \subseteq \mathcal{N}^j$ and then updates $\bm\theta^j = (\bm\theta_k^j: k \in \mathcal{K})$ for all agents at each decision epoch $j$ to minimize the accumulative loss function, which is given by
\begin{align}\label{lossFunc2}
 &   L_{(\mathrm{Deep-SARL})}\!\left(\bm\theta^j\right)
        &       &                                                                                                                                   \nonumber\\
 & = \textsf{E}_{\left(\bm\chi, (c, e), (u_k(\bm\chi, (c, e)): k \in \mathcal{K}), \bm\chi', (c', e')\right) \in \widetilde{\mathcal{N}}^j}\!
     \bigg[\sum_{k \in \mathcal{K}} \big(
        &       & \!\!\!\!\!(1 - \gamma) \cdot u_k(\bm\chi, (c, e)) + \gamma \cdot Q_k\!\left(\bm\chi', (c', e'); \bm\theta_{k, -}^j\right) -       \nonumber\\
 &      &       & \!\!\!\!\!Q_k\!\left(\bm\chi, (c, e); \bm\theta_k^j\right)\!\big)^2\bigg],
\end{align}
where $\bm\theta_{-}^j = \left(\bm\theta_{k, -}^j: k \in \mathcal{K}\right)$ are the parameters of the target DQNs of all agents at some previous decision epoch before epoch $j$.
The gradient for each agent $k \in \mathcal{K}$ can be calculated as
\begin{align}\label{grad2}
 & \nabla_{\bm\theta_k^j} L_{(\mathrm{Deep-SARL})}\!\left(\bm\theta^j\right)
        &       &                                                                                                                                   \nonumber\\
 & = \textsf{E}_{\left(\bm\chi, (c, e), (u_k(\bm\chi, (c, e)): k \in \mathcal{K}), \bm\chi', (c', e')\right) \in \widetilde{\mathcal{N}}^j}\!\big[\!\!\!\!\!
        &       & \!\!\!\!\! \big((1 - \gamma) \cdot u_k(\bm\chi, (c, e)) + \gamma \cdot Q_k\!\left(\bm\chi', (c', e'); \bm\theta_{k, -}^j\right) - \nonumber\\
 &      &       & \!\!\!\!\! Q_k\!\left(\bm\chi, (c, e); \bm\theta_k^j\right)\!\!\big)\! \cdot
                  \nabla_{\bm\theta_k^j} Q_k\!\left(\bm\chi, (c, e); \bm\theta_k^j\right)\! \big].
\end{align}
We summarize the online implementation of our proposed Deep-SARL algorithm for solving the stochastic computation offloading in a MEC system as in Algorithm \ref{algo1}.
\begin{algorithm}[t]
    \caption{Online Deep-SARL Algorithm for Stochastic Computation Task Offloading in A MEC System}
    \label{algo1}
    \begin{algorithmic}[1]
        \STATE \textbf{initialize} for $j = 1$, the replay memory $\mathcal{N}^j$ with a finite size of $N \in \mathds{N}_+$, the mini-batch $\widetilde{\mathcal{N}}^j$ with a size of $\tilde{N} < N$ for experience replay, DQNs and target DQNs with two sets $\bm\theta^j = \left(\bm\theta_k^j: k \in \mathcal{K}\right)$ and $\bm\theta_{-}^j = \left(\bm\theta_{k, -}^j: k \in \mathcal{K}\right)$ of random parameters, the initial network state $\bm\chi^j \in \mathcal{X}$, a joint control action $(c^j, e^j) \in \mathcal{Y}$ randomly with probability $\epsilon$ or $(c^j, e^j) = \arg\max_{(c, e) \in \mathcal{Y}} \sum_{k \in \mathcal{K}} Q_k(\bm\chi^j, (c, e); \bm\theta_k^j)$ with probability $1 - \epsilon$.

        \REPEAT
            \STATE After performing the selected joint control action $(c^j, e^j)$, the agents realize immediate utilities $(u_k(\bm\chi^j, (c^j, e^j)): k \in \mathcal{K})$.

            \STATE The MU observes the new network state $\bm\chi^{j + 1} \in \mathcal{X}$ at the next epoch $j + 1$, which is taken as an input to the DQNs of all agents with parameters $\bm\theta^j$, and selects a joint control action $(c^{j + 1}, e^{j + 1}) \in \mathcal{Y}$ randomly with probability $\epsilon$ or $(c^{j + 1}, e^{j + 1}) = \arg\max_{(c, e) \in \mathcal{Y}} \sum_{k \in \mathcal{K}} Q_k(\bm\chi^{j + 1}, (c, e); \bm\theta_k^j)$ with probability $1 - \epsilon$.

            \STATE The replay memory $\mathcal{N}^j$ at the mobile device is updated with the most recent transition $\mathbf{n}^j = \left(\bm\chi^j, (c^j, e^j), (u_k(\bm\chi^j, (c^j, e^j)): k \in \mathcal{K}), \bm\chi^{j + 1}, (c^{j + 1}, e^{j + 1})\right)$.

            \STATE With a randomly sampled mini-batch of transitions $\widetilde{\mathcal{N}}^j \subseteq \mathcal{N}^j$, all agents update the DQN parameters $\bm\theta^j$ using the gradient as in (\ref{grad2}).

            \STATE The target DQNs are regularly reset by $\bm\theta_{-}^{j + 1} = \bm\theta^j$, and otherwise $\bm\theta_{-}^{j + 1} = \bm\theta_{-}^j$.

            \STATE The decision epoch index is updated by $j \leftarrow j + 1$.
        \UNTIL{A predefined stopping condition is satisfied.}
    \end{algorithmic}
\end{algorithm}

\section{Numerical Experiments}
\label{simu}

In this section, we proceed to evaluate the stochastic computation offloading performance achieved from our derived online learning algorithms, namely, DARLING and Deep-SARL.

\subsection{General Setup}

Throughout the experiments, we suppose there are $B = 6$ BSs in the system connecting the MU with the MEC server.
The channel gain states between the MU and the BSs are from a common finite set $\{-11.23, -9.37, -7.8, -6.3, -4.68, -2.08\}$ (dB), the transitions of which happen across the discrete decision epochs following respective randomly generated matrices.
Each energy unit corresponds to $2 \cdot 10^{-3}$ Joule, and the number of energy units received from the wireless environment follows a Poisson arrival process with average arrival rate $\lambda_{(\mathrm{e})}$ (in units per epoch).
We set $K = 5$ for the Deep-SARL algorithm, while the $u^{(1)}\!\left(\bm\chi^j, (c^j, e^j)\right)$, $u^{(2)}\!\left(\bm\chi^j, (c^j, e^j)\right)$, $u^{(3)}\!\left(\bm\chi^j, (c^j, e^j)\right)$, $u^{(4)}\!\left(\bm\chi^j, (c^j, e^j)\right)$ and $u^{(5)}\!\left(\bm\chi^j, (c^j, e^j)\right)$ in (\ref{utilFunc}) are chosen to be the exponential functions.
Then,
\begin{align}
     u_1\!\left(\bm\chi^j, (c^j, e^j)\right) & = \omega_1 \cdot u^{(1)}\!\left(\bm\chi^j, (c^j, e^j)\right)
                                               = \omega_1 \cdot \exp\!\left\{- \min\!\left\{d^j, \delta\right\}\right\},     \\
     u_2\!\left(\bm\chi^j, (c^j, e^j)\right) & = \omega_2 \cdot u^{(2)}\!\left(\bm\chi^j, (c^j, e^j)\right)
                                               = \omega_2 \cdot \exp\!\left\{- \eta^j\right\},                               \\
     u_3\!\left(\bm\chi^j, (c^j, e^j)\right) & = \omega_3 \cdot u^{(3)}\!\left(\bm\chi^j, (c^j, e^j)\right)
                                               = \omega_3 \cdot \exp\!\left\{- \rho^j\right\},                               \\
     u_4\!\left(\bm\chi^j, (c^j, e^j)\right) & = \omega_4 \cdot u^{(4)}\!\left(\bm\chi^j, (c^j, e^j)\right)
                                               = \omega_4 \cdot \exp\!\left\{- \varphi^j\right\},                            \\
     u_5\!\left(\bm\chi^j, (c^j, e^j)\right) & = \omega_5 \cdot u^{(5)}\!\left(\bm\chi^j, (c^j, e^j)\right)
                                               = \omega_5 \cdot \exp\!\left\{- \phi^j\right\}.
\end{align}
Based on the works \cite{Chen18} and \cite{He16}, we use a single hidden layer consisting of 200 neurons\footnote{The tradeoff between a time demanding training process and an improvement in performance with a deeper and/or wider neural network is still an open problem \cite{CCLP18}.} for the design of a DQN in DARLING algorithm and select tanh as the activation function \cite{Jarr09} and Adam as the optimizer.
The same number of 200 neurons are employed to design the single-layer DQNs of all agents in Deep-SARL and hence for each DQN of an agent, there are in total 40 neurons.
Both the DARLING and the Deep-SARL algorithms are experimented in TensorFlow \cite{Abad16}.
Other parameter values used in experiments are listed in Table \ref{tabl2}.

\begin{table}[t]
  \caption{Parameter values in experiments.}\label{tabl2}
        \begin{center}
        \begin{tabular}{c|c}
              \hline
              Parameter                                                     & Value                                                             \\\hline
              Replay memory capacities        $M$, $N$                      & $5000$, $5000$                                                    \\\hline
              Mini-batch sizes                $\tilde{M}$, $\tilde{N}$      & $200$, $200$                                                      \\\hline
              Decision epoch duration         $\delta$                      & $5 \cdot 10^{-3}$ second                                          \\\hline
              Channel bandwidth               $W$                           & $0.6$ MHz                                                         \\\hline
              Noise power                     $I$                           & $1.5 \cdot 10^{-8}$ Watt                                          \\\hline
              Input data size                 $\mu$                         & $10^4$ bits                                                       \\\hline
              CPU cycles                      $\nu$                         & $7.375 \cdot 10^6$                                                \\\hline
              Maximum CPU-cycle frequency     $f_{(\mathrm{CPU})}^{(\max)}$ & $2 \times 10^9$ Hz                                                \\\hline
              Maximum transmit power          $p_{(\mathrm{tr})}^{(\max)}$  & $2$ Watt                                                          \\\hline
              Handover delay                  $\zeta$                       & $2 \times 10^{-3}$ second                                         \\\hline
              MEC service price               $\xi$                         & $1$                                                               \\\hline
              Weights $\omega_1$, $\omega_2$, $\omega_3$,
                      $\omega_4$, $\omega_5$                                & $3$, $9$, $5$, $2$, $1$                                           \\\hline
              Maximum task queue length       $q_{(\mathrm{t})}^{(\max)}$   & $4$ tasks                                                         \\\hline
              Maximum energy queue length     $q_{(\mathrm{e})}^{(\max)}$   & $4$ units                                                         \\\hline
              Discount factor                 $\gamma$                      & $0.9$                                                             \\\hline
              Exploration probability         $\epsilon$                    & $0.01$                                                            \\
              \hline
        \end{tabular}
    \end{center}
\end{table}

For the purpose of performance comparisons, we simulate three baselines as well, namely,
\begin{enumerate}
  \item \emph{Mobile Execution} -- The MU processes all scheduled computation task at the mobile device with maximum possible energy, that is, at each decision epoch $j$, $c^j = 0$ and
      \begin{align}
        e^j = \left\{\!\!
        \begin{array}{ll}
          \min\!\left\{q_{(\mathrm{e})}^j, \left\lfloor \left(f_{(\mathrm{CPU})}^{(\max)}\right)^3 \cdot \tau\right\rfloor\right\},
                                                        & \mbox{if } q_{(\mathrm{e})}^j > 0; \\
          0,                                            & \mbox{if } q_{(\mathrm{e})}^j = 0,
        \end{array}
        \right.
      \end{align}
      where the allocation of energy units takes into account the maximum CPU-cycle frequency $f_{(\mathrm{CPU})}^{(\max)}$ and $\lfloor \cdot \rfloor$ means the floor function.
  \item \emph{Server Execution} -- According to Lemma 2, with maximum possible energy units in the energy queue that satisfy the maximum transmit power constraint, the MU always selects a BS that achieves the minimum task execution delay to offload the input data of a scheduled computation task for MEC server execution.
  \item \emph{Greedy Execution} -- At each decision epoch, the MU decides to execute a computation task at its own mobile device or offload it to the MEC server for processing with the aim of minimizing the immediate task execution delay.
\end{enumerate}

\subsection{Experiment Results}

We carry out experiments under various settings to validate the proposed studies in this paper.

\subsubsection{Experiment 1 -- Convergence performance}

\begin{figure}[t]
    \centering
    \includegraphics[width=19pc]{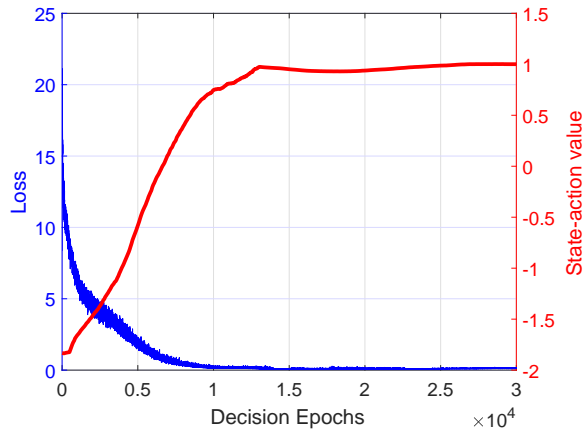}
    \caption{Illustration for the convergence property of DARLING and Deep-SARL.}
    \label{conver}
\end{figure}

Our goal in this experiment is to validate the convergence property of our proposed algorithms, namely, DARLING and Deep-SARL, for stochastic computation offloading in the considered MEC system.
We set the task arrival probability and the average energy arrival rate to be $\lambda_{(\mathrm{t})} = 0.5$ and $\lambda_{(\mathrm{e})} = 0.8$ units per epoch, respectively.
Without loss of the generality, we plot the simulated variations in $Q\!\left(\bm\chi, (c, e) ; \bm\theta^j\right)$ (where $\bm\chi = (2, 2, 2, (-6.3, -6.3, -4.68, -7.8, -6.3, -6.3))$, $(c, e) = (2, 4)$ and $j \in \mathds{N}_+$) for the DARLING algorithm as well as the loss function defined by (\ref{lossFunc2}) for the Deep-SARL algorithm versus the decision epochs in Fig. \ref{conver}, which reveals that the convergence behaviours of both DARLING and Deep-SARL are similar.
The two algorithms converge at a reasonable speed, for which around $1.5 \times 10^4$ decision epochs are needed.

\subsubsection{Experiment 2 -- Performance under various $\lambda_{(\mathrm{t})}$}

\begin{figure}
  \centering
  \subfigure[Average utility per epoch.]{\label{perf01: simu02_01}\includegraphics[width=19pc]{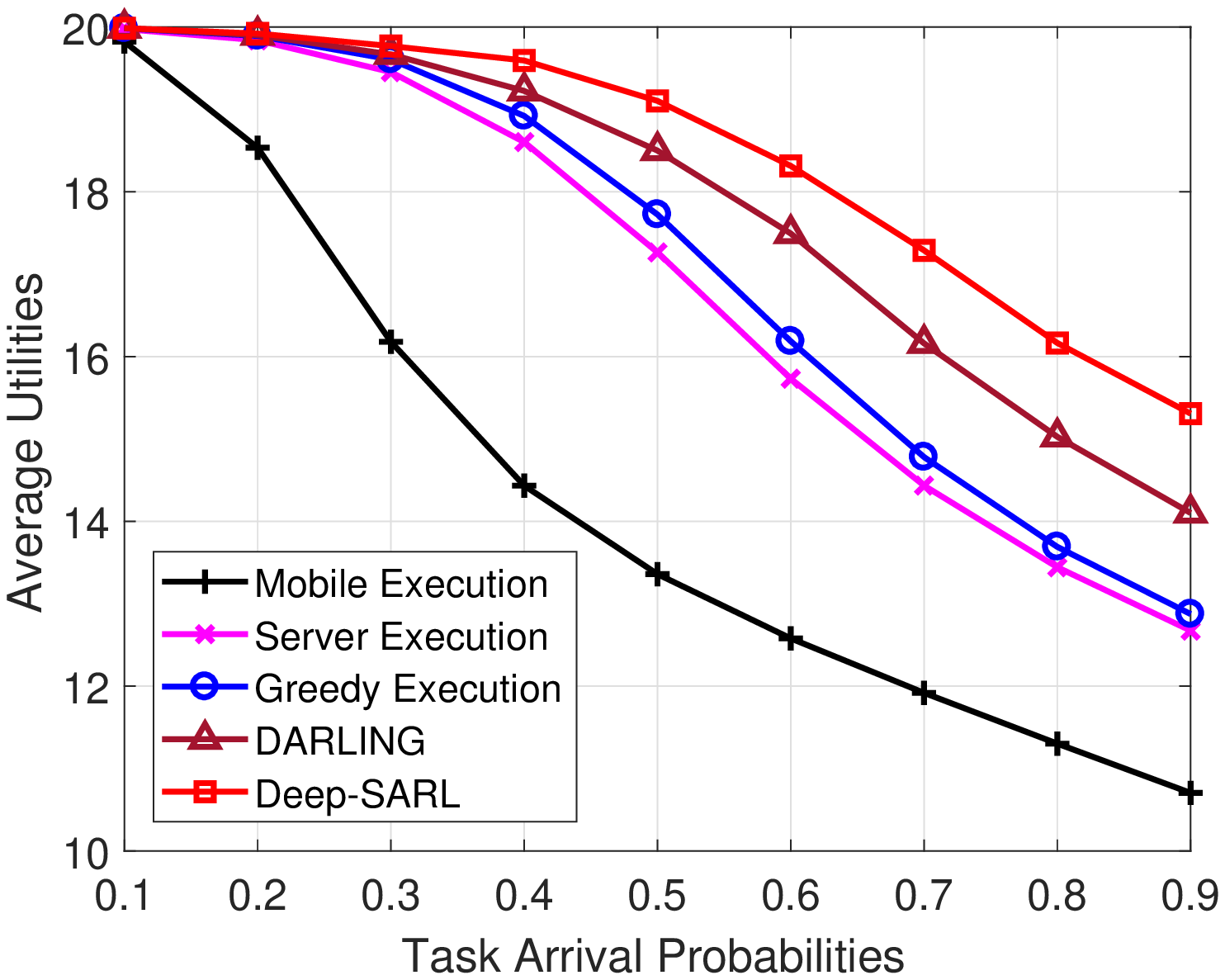}}
  \subfigure[Average execution delay per epoch.]{\label{perf01: simu02_03}\includegraphics[width=19pc]{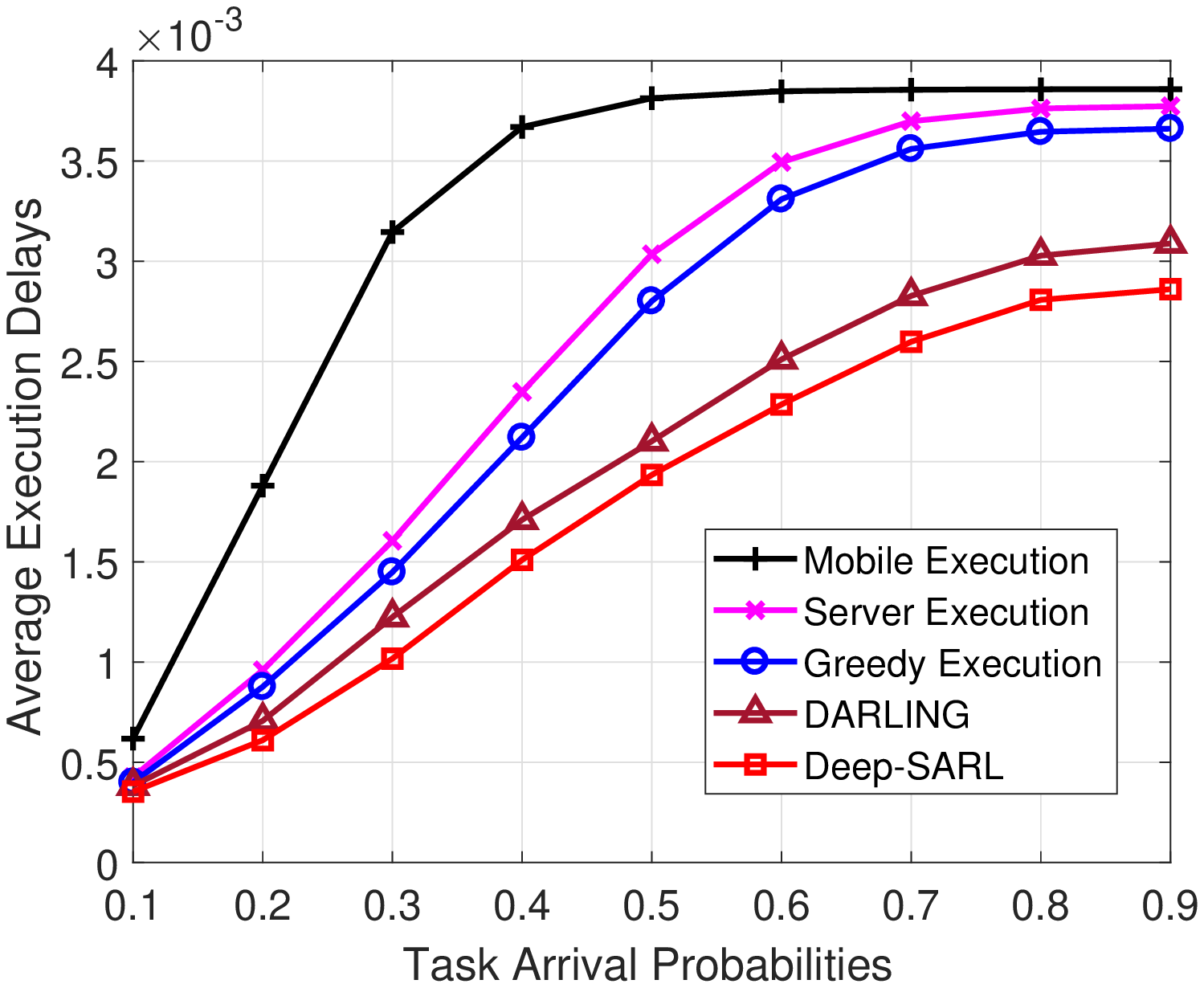}}
  \subfigure[Average task drops per epoch.]{\label{perf01: simu02_02}\includegraphics[width=19pc]{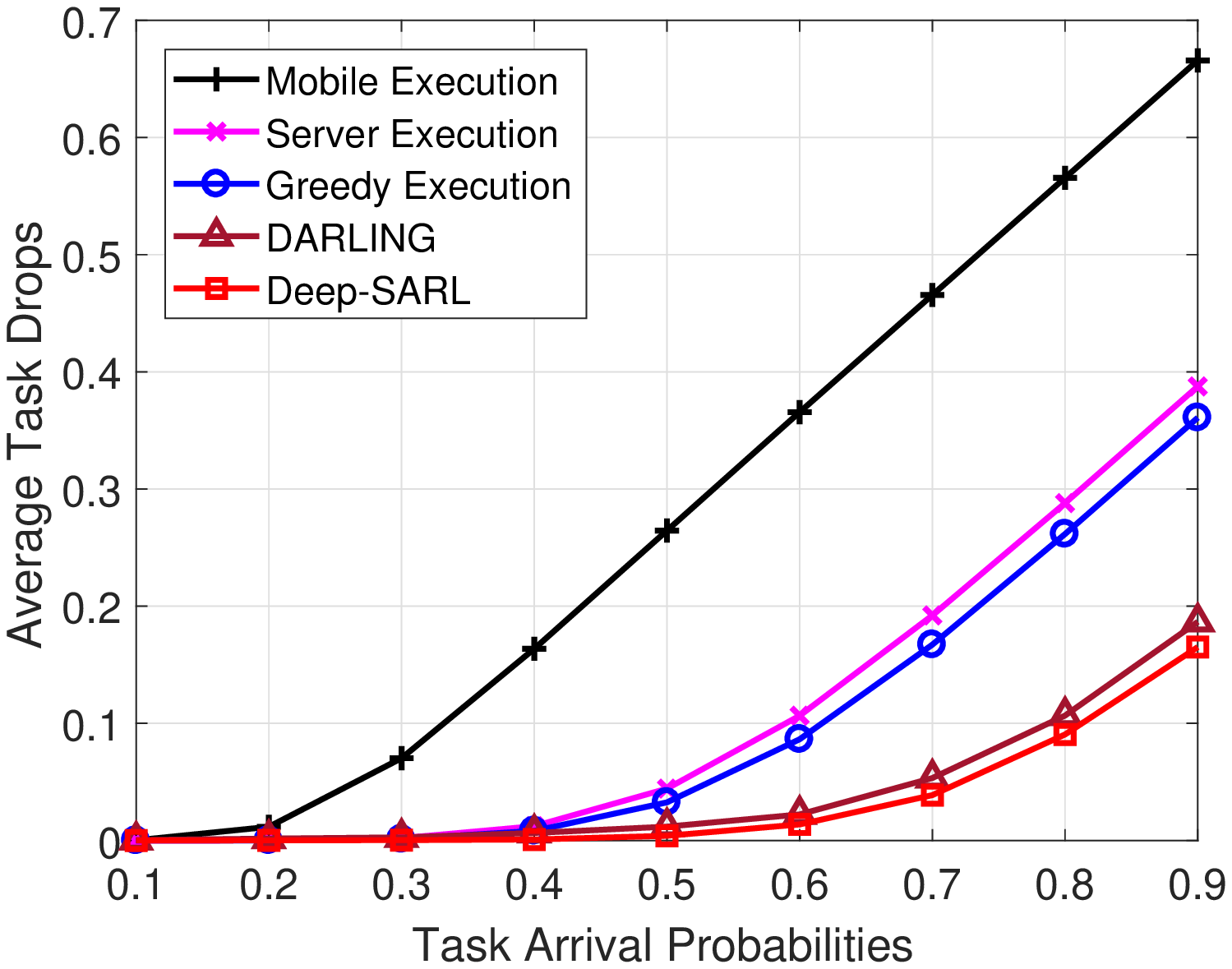}}
  \subfigure[Average task queuing delay per epoch.]{\label{perf01: simu02_04}\includegraphics[width=19pc]{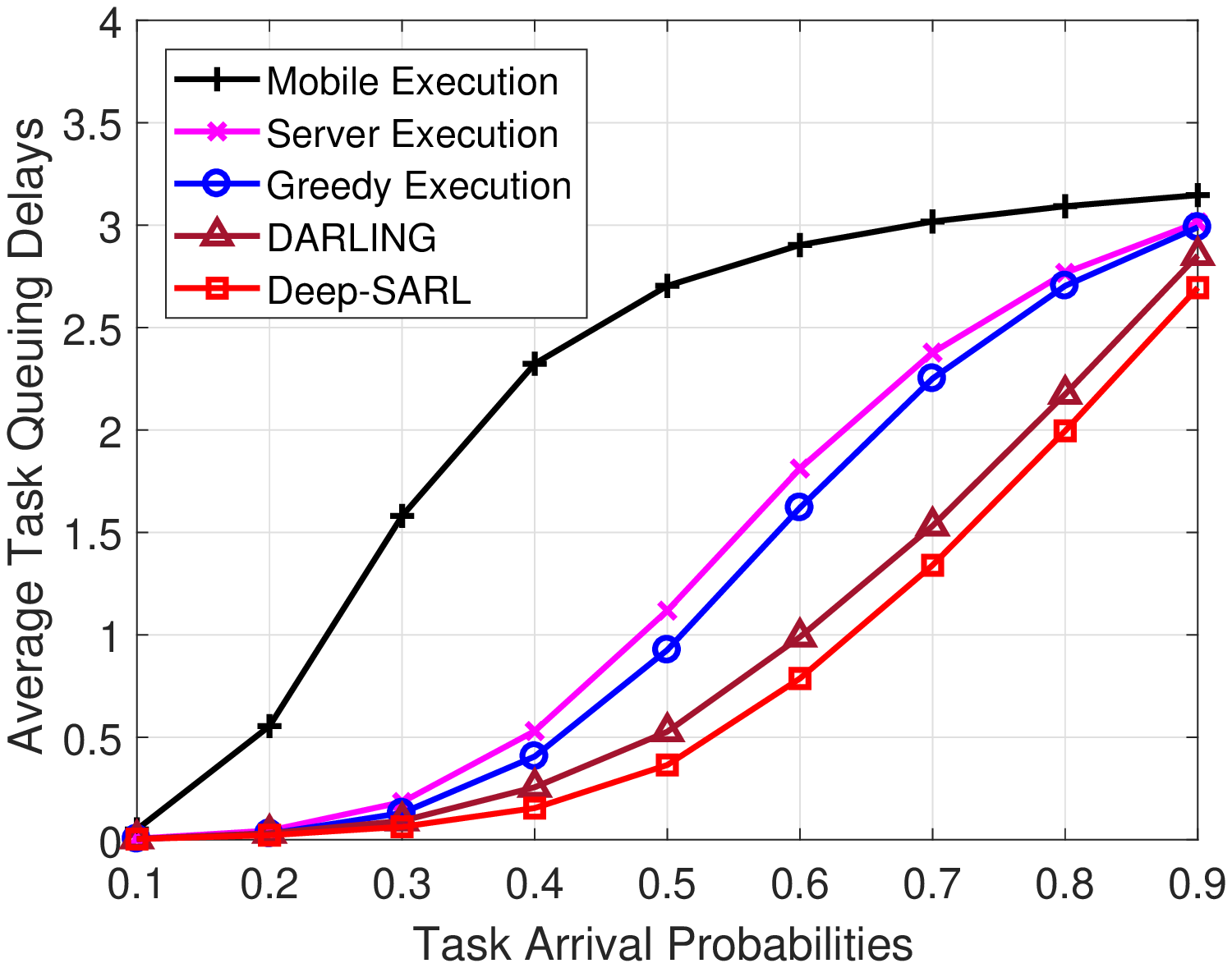}}
  \subfigure[Average MEC service payment per epoch.]{\label{perf01: simu02_05}\includegraphics[width=19pc]{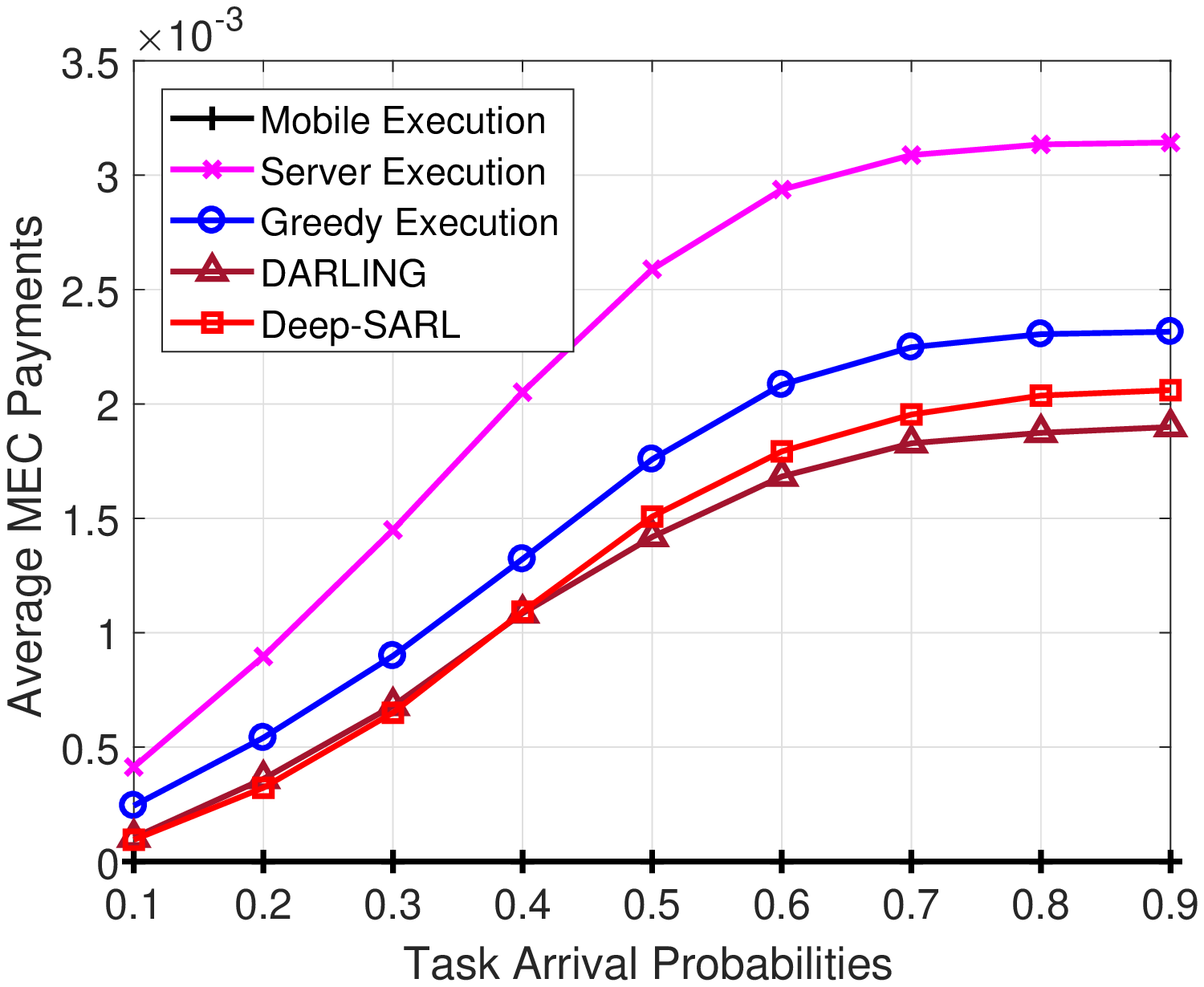}}
  \subfigure[Average task execution failure penalty per epoch.]{\label{perf01: simu02_06}\includegraphics[width=19pc]{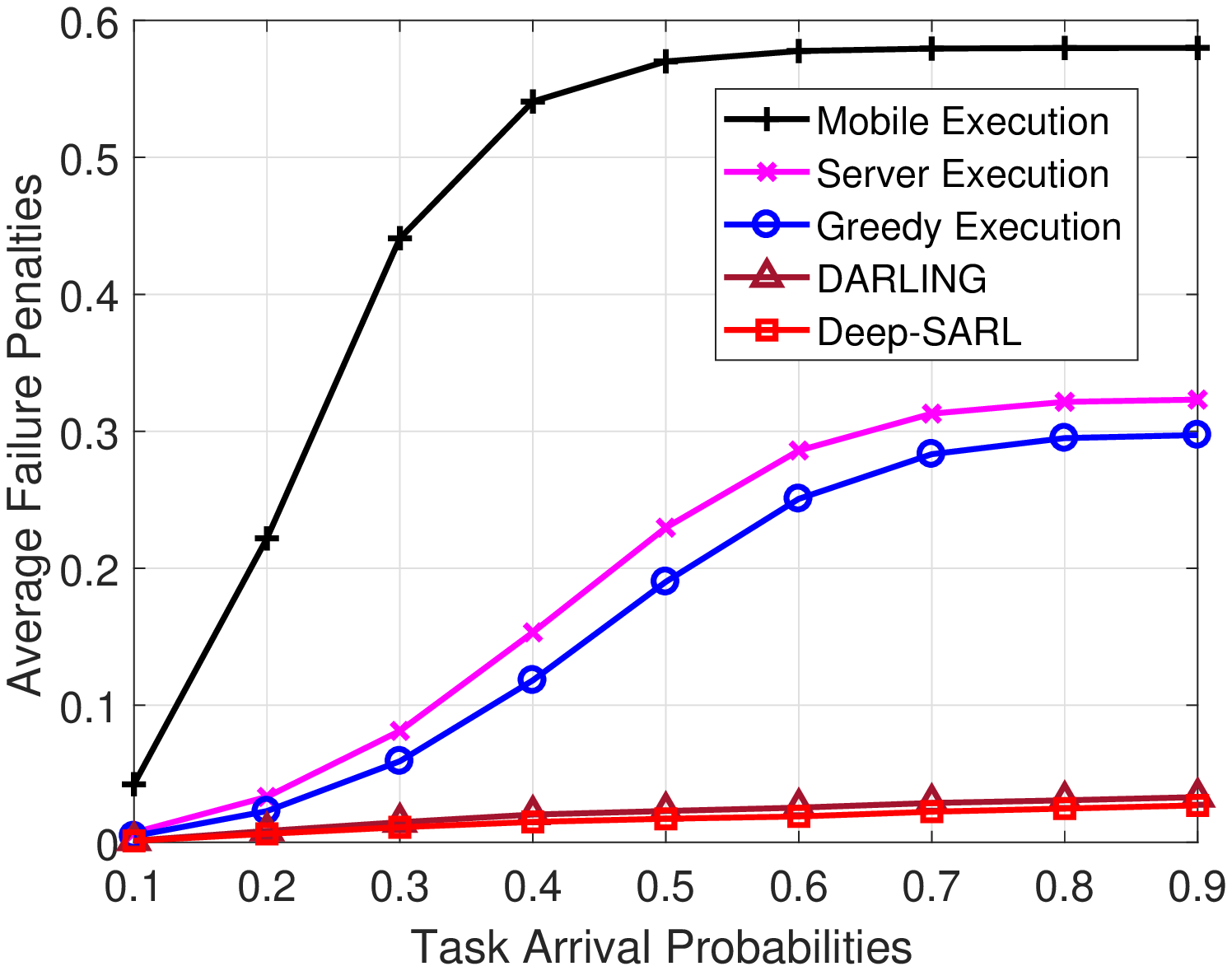}}
  \caption{Average computation offloading performance versus task arrival probabilities.}\label{perf02}
\end{figure}

In this experiment, we try to demonstrate the average computation offloading performance per epoch in terms of the average utility, the average task execution delay, the average task drops, the average task queuing delay, the average MEC service payment and the average task failure penalty under different computation task arrival probability settings.
We choose for the average energy unit arrival rate as $\lambda_{(\mathrm{e})} = 1.6$ units per epoch.
The results are exhibited in Fig. \ref{perf02}.
Fig. \ref{perf02} (a) illustrates the average utility performance when the MU implements DARLING and Deep-SARL.
Figs. \ref{perf02} (b)--(f) illustrate the average task execution delay, the average task drops, the average task queuing delay, the average MEC service payment and the average task failure penalty.

Each plot compares the performance of the DARLING and the Deep-SARL to the three baseline computation offloading schemes.
From Fig. \ref{perf02}, it can be observed that both the proposed schemes achieve a significant gain in average utility.
Similar observations can be made from the curves in other plots, though the average MEC service payment per epoch from Deep-SARL is a bit higher than that from DARLING.
This can be explained by the reason that with the network settings by increasing the task arrival probability in this experiment, more tasks are scheduled for execution at the MEC server using the Deep-SARL algorithm.
As the computation task arrival probability increases, the average utility performances decrease due to the increases in average task execution delay, average task drops, average task queuing delay, average MEC service payment and average task failure penalty.
Since there are not enough energy units in the energy queue during one decision epoch on average, in order to avoid task drops and task failure penalty, only a portion of the queued energy units are allocated for processing a scheduled computation task, hence leading to more queued tasks, i.e., an increased average task queueing delay.
The utility performance from Deep-SARL outperforms that from DARLING.
This indicates that by combining a deep neural network and the $Q$-function decomposition technique, the original stochastic computation offloading problem becomes simplified, hence performance improvement can be expected from approximating the state-action $Q$-functions of all agents with the same number of neurons.

\subsubsection{Experiment 3 -- Performance with changing $\lambda_{(\mathrm{e})}$}

\begin{figure}
  \centering
  \subfigure[Average utility per epoch.]{\label{simu03_01}\includegraphics[width=19pc]{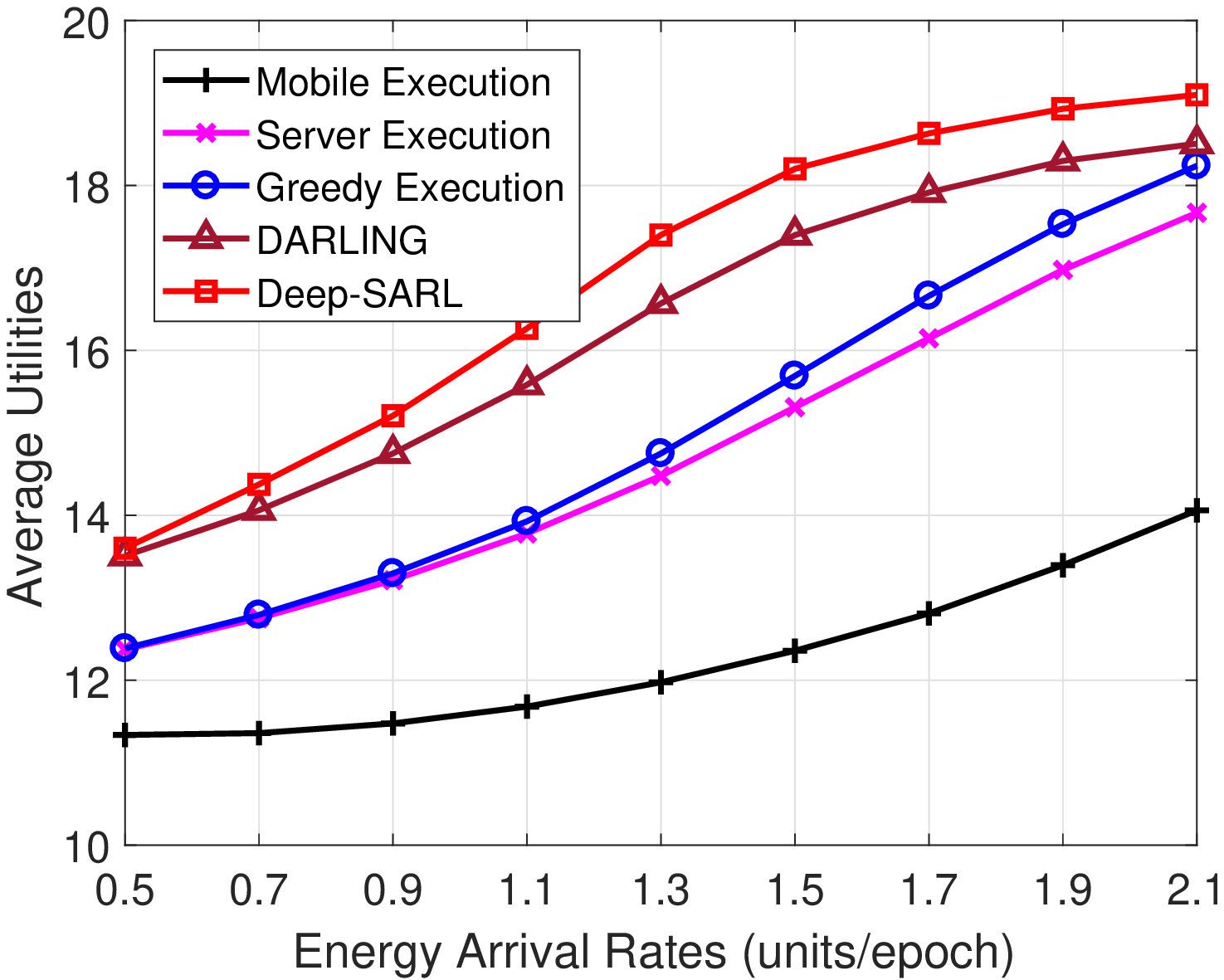}}
  \subfigure[Average execution delay per epoch.]{\label{simu03_03}\includegraphics[width=19pc]{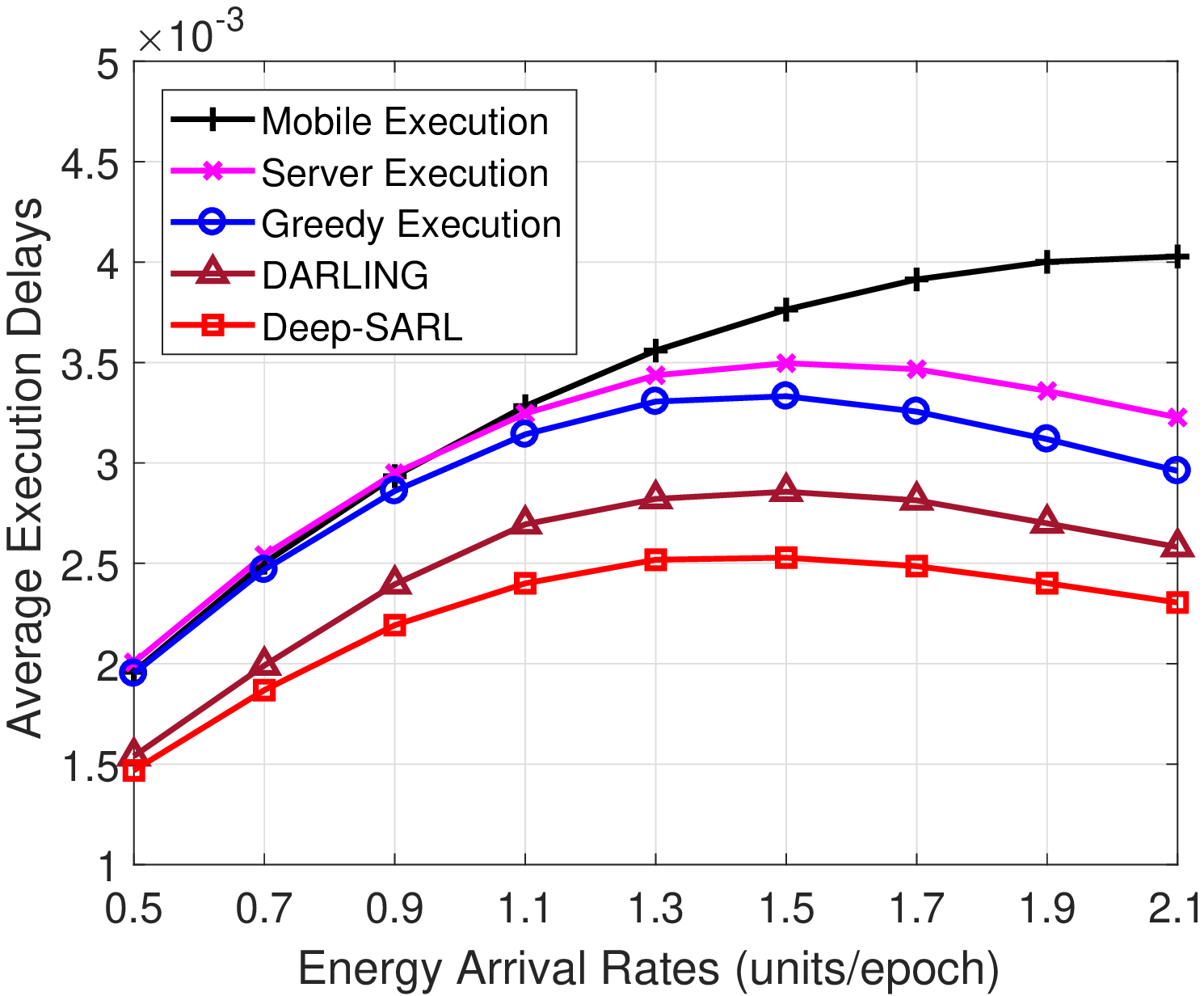}}
  \subfigure[Average task drops per epoch.]{\label{simu03_02}\includegraphics[width=19pc]{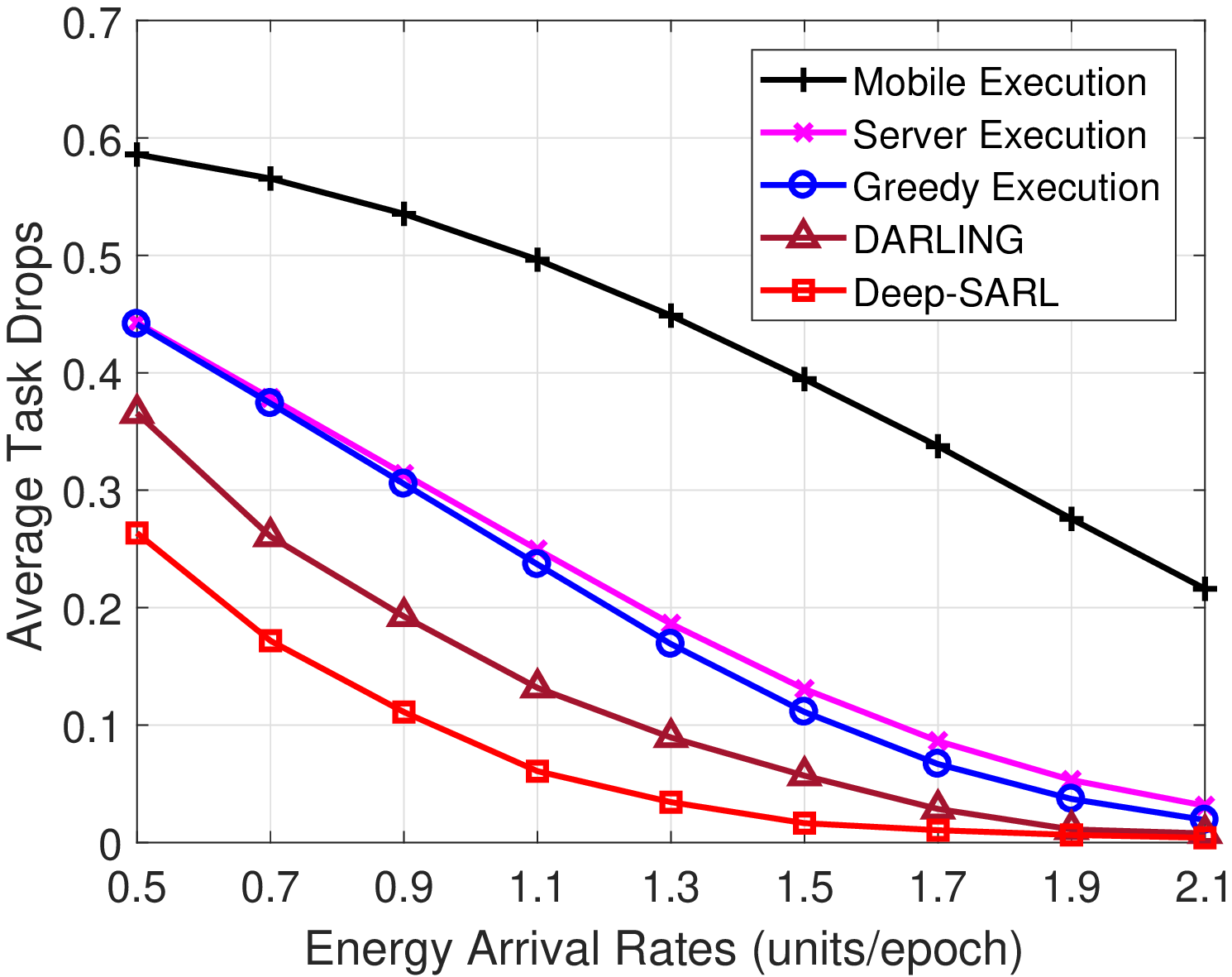}}
  \subfigure[Average task queuing delay per epoch.]{\label{simu03_04}\includegraphics[width=19pc]{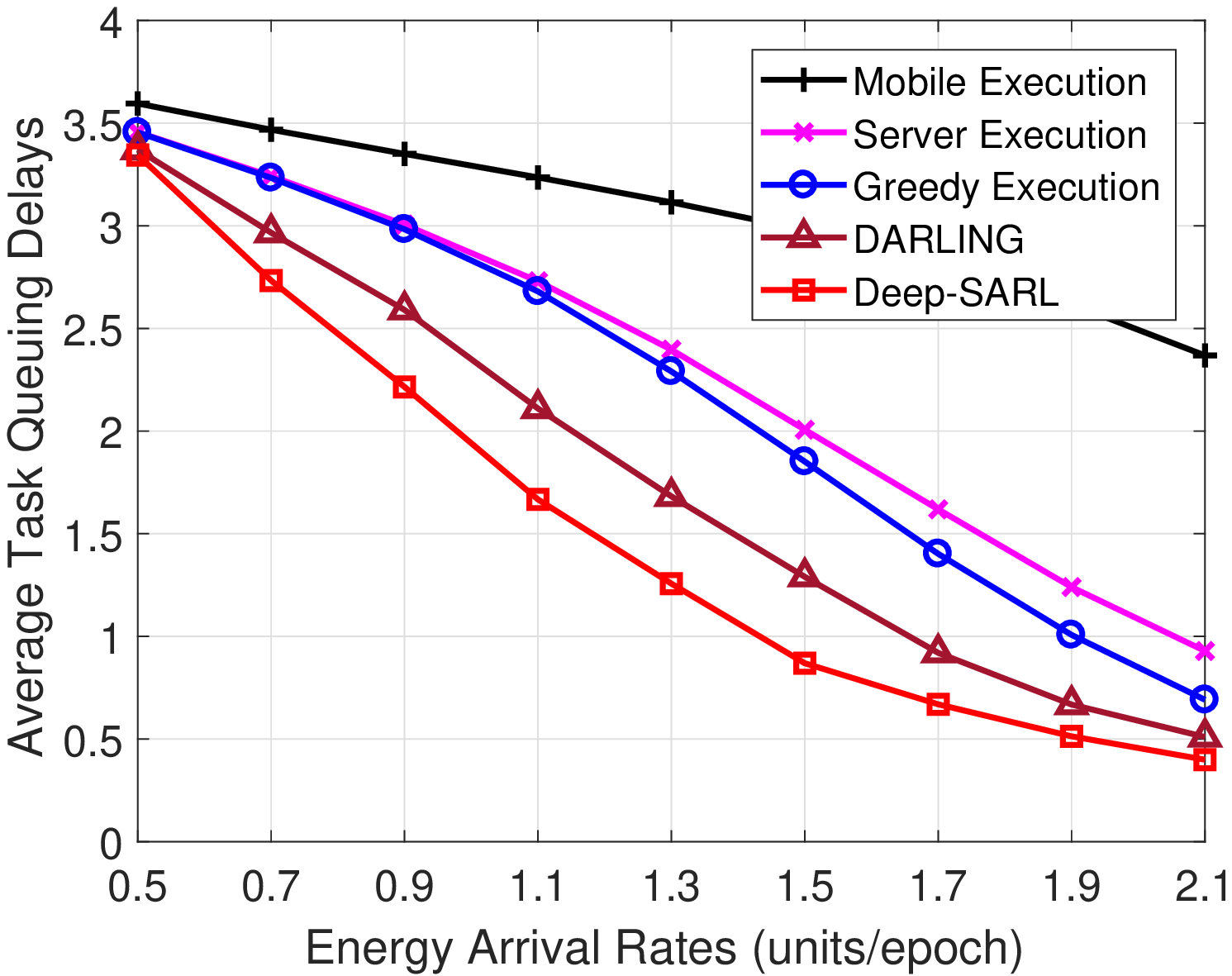}}
  \subfigure[Average MEC service payment per epoch.]{\label{simu03_05}\includegraphics[width=19pc]{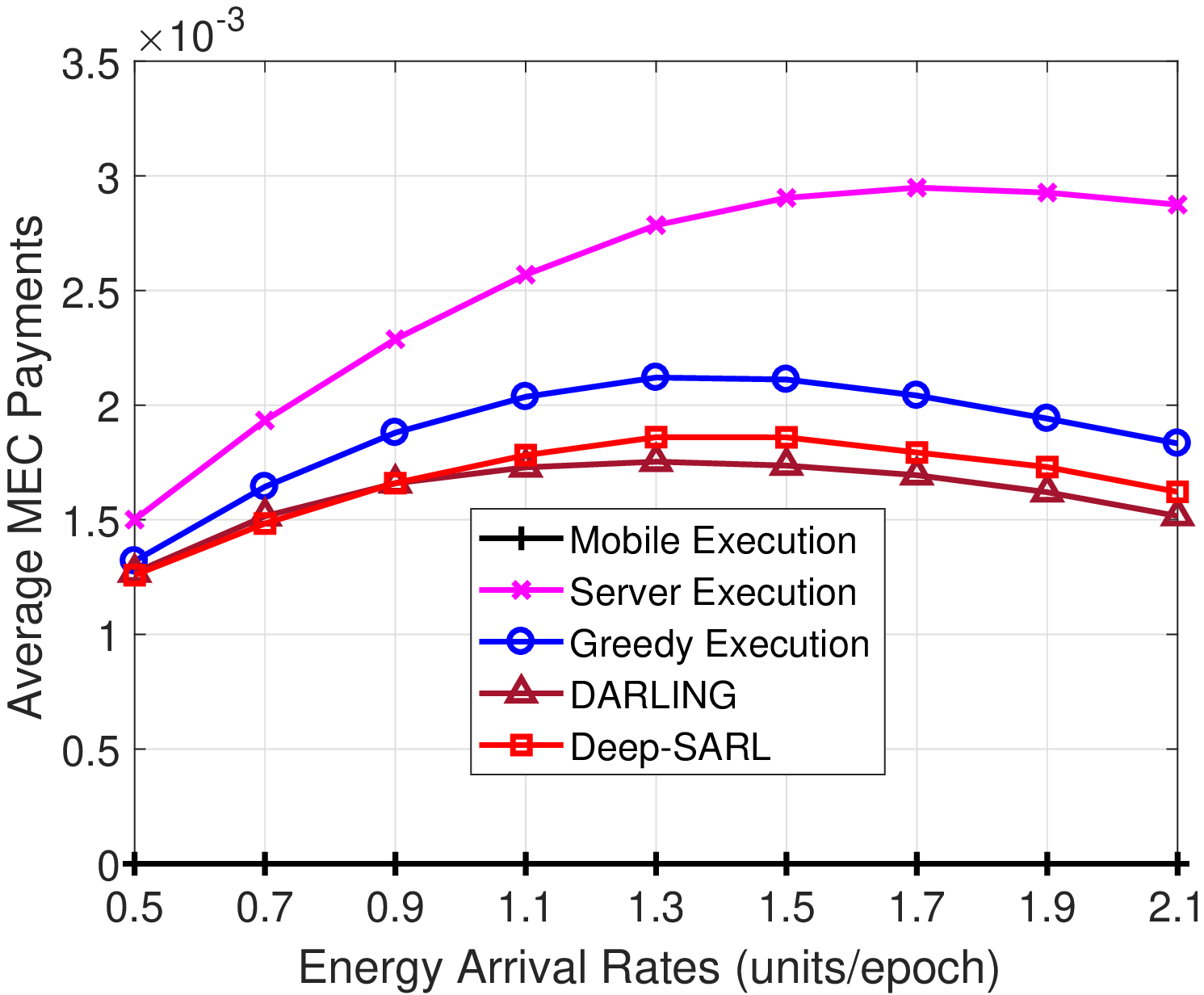}}
  \subfigure[Average task execution failure penalty per epoch.]{\label{simu03_06}\includegraphics[width=19pc]{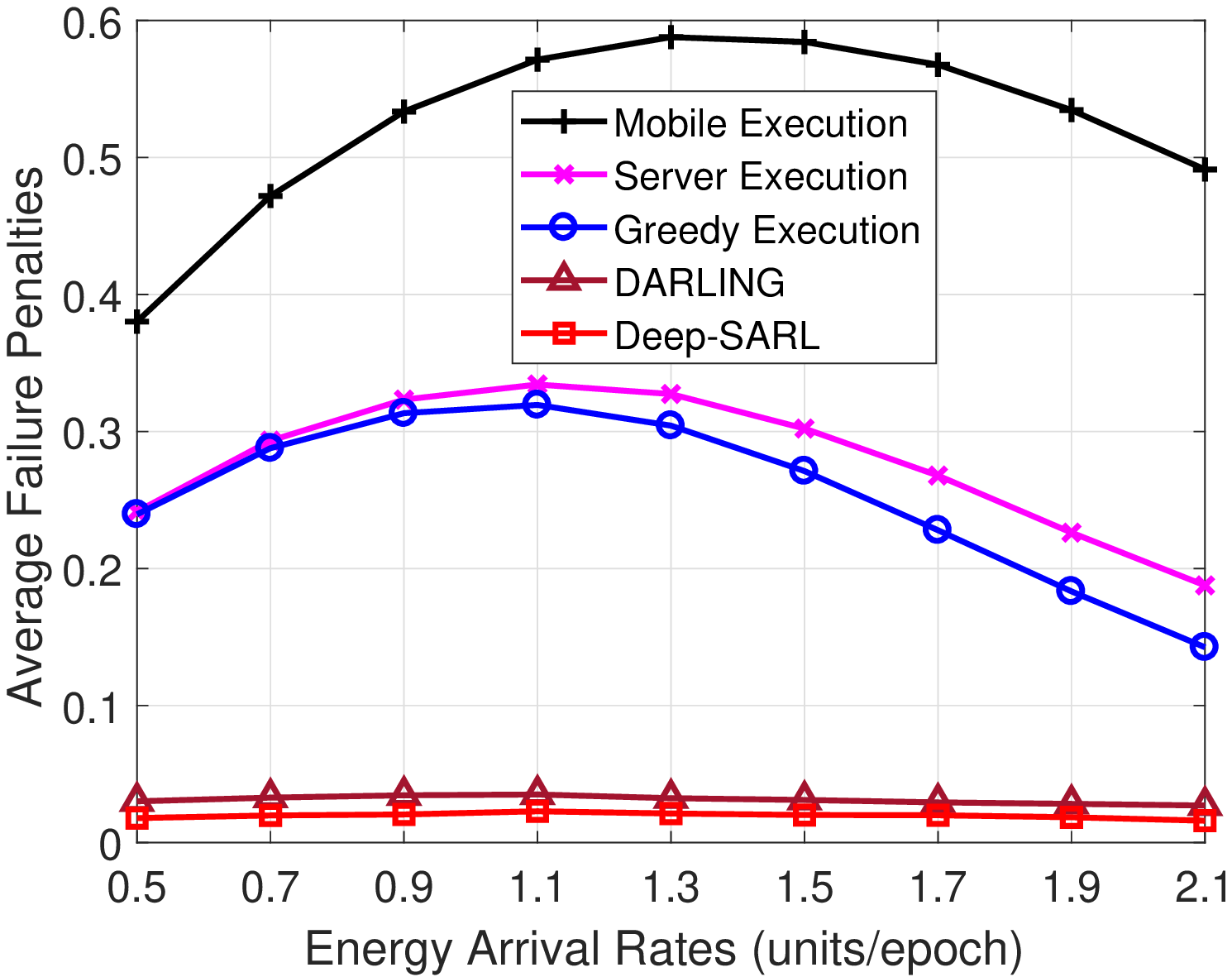}}
  \caption{Average computation offloading performance versus average energy unit arrival rates.}\label{perf03}
\end{figure}

We do the third experiment to simulate the average computation offloading performance per epoch achieved from the derived DARLING and Deep-SARL algorithms and other three baselines versus the average energy unit arrival rates.
The computation task arrival probability in this experiment is set to be $\lambda_{(\mathrm{t})} = 0.6$.
The per epoch average utility, average task execution delay, average task drops, average task queuing delay, average MEC service payment and average task failure penalty across the entire learning period are depicted in Fig. \ref{perf03}.
We can clearly see from Fig. \ref{perf03} that as the available energy units increase, the overall computation offloading performance improves.
However, as the energy unit arrival rate increases, the average task execution delay, the average MEC service payment\footnote{It's easy to see that the mobile execution scheme does not use MEC service, hence no MEC service payment needs to be made.} and the average task failure penalty first increase but then decrease.
The increasing number of queued energy units provides more opportunities to execute a computation task during each decision epoch, and at the same time, increases the possibility of failing to execute a task.
When the average number of energy units in the energy queue increases to a sufficiently large value, enough energy units can be allocated to each scheduled computation task, due to which the task execution delay, the MEC service payment as well as the possibility of task computation failures decrease.

\section{Conclusions}
\label{conc}

In this paper, we put our emphasis to investigate the design of a stochastic computation offloading policy for a representative MU in an ultra dense sliced RAN by taking into account the dynamics generated from the time-varying channel qualities between the MU and the BSs, energy units received from the wireless environment as well as computation task arrivals.
The problem of stochastic computation offloading is formulated as a MDP, for which we propose two double DQN-based online strategic computation offloading algorithms, namely, DARLING and Deep-SARL.
Both learning algorithms survive the curse of high dimensionality in state space and need no a priori information of dynamics statistics.
We find from numerical experiments that compared to three baselines, our derived algorithms can achieve much better long-term utility performance, which indicates an optimal tradeoff among the computation task execution delay, the task drops, the task queuing delay, the MEC service payment and the task failure penalty.
Moreover, the Deep-SARL algorithm outperforms the DARLING algorithm by taking the advantage of the additive utility function structure.

\end{document}